\documentclass{article}

\usepackage{arxiv}

\usepackage[utf8]{inputenc} 
\usepackage[T1]{fontenc}    
\usepackage{hyperref}       
\usepackage{url}            
\usepackage{booktabs}       
\usepackage{amsfonts}       
\usepackage{nicefrac}       
\usepackage{microtype}      
\usepackage{amsmath} 
\usepackage{cleveref}       
\usepackage{graphicx}
\usepackage{cite}
\usepackage{doi}
\usepackage{multirow}
\usepackage[dvipsnames,table]{xcolor}
\usepackage{bbding}
\usepackage{authblk}
\usepackage{longtable}
\usepackage{setspace}
\usepackage{tikz}
\usepackage{enumitem}
\usepackage{listings}
\usepackage[most]{tcolorbox}
\usepackage[title]{appendix}
\usepackage{array}
\usepackage{colortbl}
\usepackage{xcolor}
\usepackage{booktabs}
\usepackage{multirow}
\usepackage{pifont}
\usepackage{amsthm}

\newcommand{\testpassed}{{\color{PineGreen}\ding{51}}}
\newcommand{\testfailed}{{\color{BrickRed}\ding{55}}}

\lstdefinelanguage{my-yaml}{
  keywords={type, name, llm_config, system_message,children,tools,model,api_key,base_url,python_path,description,verilog_code,filename,parameters},
  keywordstyle=\color{blue}\bfseries,
  moredelim=[is][commentstyle]{||}{££}, 
  identifierstyle=\color{black},
  sensitive=false,
  comment=[l]{\#},
  commentstyle=\color{olive}\ttfamily,
  stringstyle=\color{orange}\ttfamily,
  morestring=[b]',
  morestring=[b]"
}

\definecolor{mygreen}{rgb}{0,0.6,0}
\definecolor{mygray}{rgb}{0.5,0.5,0.5}
\definecolor{mymauve}{rgb}{0.58,0,0.82}
\definecolor{mathematiciancolor}{RGB}{230,247,255}
\definecolor{reviewercolor}{RGB}{240,255,240}
\definecolor{supervisorcolor}{RGB}{255,245,230}
\definecolor{usercolor}{RGB}{255,240,245}
\definecolor{userborder}{RGB}{128,0,0}

\newtcolorbox{userbox}{
    colback=usercolor,
    colframe=userborder,
    arc=7pt,
    outer arc=7pt,
    boxrule=1pt,
    left=5pt,right=5pt,top=5pt,bottom=5pt,
    breakable
}

\newtcolorbox{mathematicianbox}{
    colback=mathematiciancolor,
    colframe=blue!75!black,
    arc=7pt,
    outer arc=7pt,
    boxrule=1pt,
    left=5pt,right=5pt,top=5pt,bottom=5pt,
    breakable
}

\newtcolorbox{reviewerbox}{
    colback=reviewercolor,
    colframe=green!75!black,
    arc=7pt,
    outer arc=7pt,
    boxrule=1pt,
    left=5pt,right=5pt,top=5pt,bottom=5pt,
    breakable
}

\newtcolorbox{supervisorbox}{
    colback=supervisorcolor,
    colframe=orange!75!black,
    arc=7pt,
    outer arc=7pt,
    boxrule=1pt,
    left=5pt,right=5pt,top=5pt,bottom=5pt,
    breakable
}

\graphicspath{{images/}}
\hypersetup{
    colorlinks,
    linkcolor={blue!90!black},
    citecolor={blue!90!black},
    urlcolor={blue!90!black}
}

\title{Nexus: A Lightweight and Scalable Multi-Agent Framework for Complex Tasks Automation}

\date{}

\newbox{\orcid}\sbox{\orcid}{\includegraphics[scale=0.06]{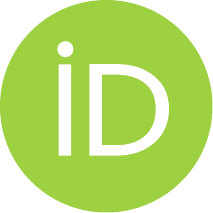}}

\author{%
    \bf Humza Sami\textsuperscript{1}, Mubashir ul Islam\textsuperscript{1}, Samy Charas\textsuperscript{1}\\
    \bf Asav Gandhi\textsuperscript{1}, Pierre-Emmanuel Gaillardon\textsuperscript{1,2}, Valerio Tenace\textsuperscript{1}\thanks{Corresponding author: \texttt{valerio@primis.ai}}%
    \\
    \vspace{1em}
    \textsuperscript{1}PrimisAI, Los Gatos, CA, USA\\
    \textsuperscript{2}University of Utah, Salt Lake City, UT, USA
}

\renewcommand{\headeright}{}
\renewcommand{\undertitle}{}
\renewcommand{\shorttitle}{Nexus: A Lightweight and Scalable Multi-Agent Framework}

\hypersetup{
pdftitle={Nexus: A Lightweight and Scalable Multi-Agent Framework},
pdfsubject={cs.AI, cs.MA},
pdfauthor={},
pdfkeywords={Large Language Models, Multi-Agent Systems, Generative AI},
}

\begin{document}
\maketitle

\setcounter{footnote}{0}

\newtheoremstyle{basic}
    {0pt}{0pt}{\normalfont}{0pt}
    {}{\;}{0.25em}
    {{\bfseries\color{blue!50!cyan}\thmname{#1}~\thmnumber{\textup{#2}}.}
        \thmnote{\normalfont\color{black}~(#3)}}

\newtheoremstyle{thinkingstyle}
        {0pt}{0pt}{\normalfont}{0pt}%
        {}{\;}{0.25em}%
        {%
          \if\relax\detokenize{#3}\relax
            {\bfseries\color{red!50!yellow}\thmname{#1}~\thmnumber{\textup{#2}}.}%
          \else
            {\bfseries\color{red!50!yellow}#3}%
          \fi%
        }

\theoremstyle{basic}
\newtheorem{problem}{Problem}

\tcolorboxenvironment{problem}{
    enhanced jigsaw, pad at break*=1mm, breakable,
    left=4mm, right=4mm, top=1mm, bottom=1mm,
    colback=blue!50!cyan!10, boxrule=0pt, frame hidden,
    borderline west={0.5mm}{0mm}{blue!50!cyan}, arc=.5mm
}

\theoremstyle{thinkingstyle}
\newtheorem{thinking}{Thinking Process}

\tcolorboxenvironment{thinking}{
    enhanced jigsaw, pad at break*=1mm, breakable,
    left=4mm, right=4mm, top=1mm, bottom=1mm,
    colback=red!50!yellow!10, boxrule=0pt, frame hidden,
    borderline west={0.5mm}{0mm}{red!50!yellow}, arc=.5mm
}

\begin{spacing}{.99}
\vspace{-2em}
\begin{abstract}
Recent advancements in {\em Large Language Models} (LLMs) have substantially evolved {\em Multi-Agent Systems} (MASs) capabilities, enabling systems that not only automate tasks but also leverage {\em near-human} reasoning capabilities. To achieve this, LLM-based MASs need to be built around two critical principles: ($i$) a robust architecture that fully exploits LLM potential for specific tasks\textemdash or related task sets\textemdash and ($ii$) an effective methodology for equipping LLMs with the necessary capabilities to perform tasks and manage information efficiently. It goes without saying that {\em a priori} architectural designs can limit the scalability and domain adaptability of a given MAS. Furthermore, complex MAS architectures may rely on overcomplex code implementations, thus making reusability of the same codebase to other scenarios near to impossible.

To address these challenges, in this paper we introduce Nexus: a lightweight Python framework designed to easily build and manage LLM-based MASs. Nexus introduces several innovations, with key contributions summarized as follows: ($i$) a {\bf flexible multi-supervisor hierarchy}: Nexus supports hierarchical architectures with a global supervisor orchestrating the overall workflow and delegating subsets of tasks to specialized supervisors, each controlling a smaller group of agents. This divide-and-conquer approach enables efficient handling of highly complex tasks and improves scalability; ($ii$) a {\bf simplified workflow design}: users can design custom architectures and workflows through YAML files, thus drastically reducing, if not completely eliminating, the need for programming expertise; and ($iii$) {\bf easy installation and open-source flexibility}: Nexus can be installed via \texttt{pip} and is distributed under a permissive open-source license, allowing users to freely modify and extend its capabilities\footnote{Source code available at \url{https://github.com/PrimisAI/nexus}}.

Experimental results demonstrate that architectures built with Nexus exhibit state-of-the-art performance across diverse domains. In coding tasks, Nexus-driven MASs achieve a 99\% pass rate on HumanEval and a flawless 100\% on VerilogEval-Human, outperforming cutting-edge reasoning language models such as o3-mini and DeepSeek-R1. Moreover, these architectures display robust proficiency in complex reasoning and mathematical problem solving, achieving correct solutions for all randomly selected problems from the MATH dataset. In the realm of multi-objective optimization, Nexus-based architectures successfully address challenging timing closure tasks on designs from the VTR benchmark suite, while guaranteeing, on average, a power saving of nearly 30\%.

\end{abstract}

\keywords{Large Language Models \and Multi-Agent Systems \and Generative AI}

\section{Introduction}
Since their inception in the 1980s, {\em Multi-Agent Systems} (MASs) have become foundational in {\em Distributed Artificial Intelligence}, enabling the decomposition of complex tasks into smaller, more manageable components executed by autonomous agents~\cite{vlassis2022concise,dorri2018multi}. These agents draw on historical knowledge, interactions with other agents, and environmental cues to make decisions and act autonomously. This built-in autonomy and flexibility distinguish MASs from traditional distributed problem-solving systems, enhancing their ability to operate effectively in dynamic and uncertain environments. Consequently, MASs have been widely applied in fields as diverse as robotic control~\cite{czimmermann2021autonomous,cena2013cooperative,verma2021multi}, traffic management~\cite{balaji2007multi,hamidi2018approach}, smart grids~\cite{shobole2021multiagent,merabet2014applications}, network security~\cite{gorodetski2003multi,herrero2009multiagent,gorodetski2002multi}, and many others~\cite{dorri2018multi}.

Despite their versatility, conventional MAS architectures have historically relied on predefined rules and heuristic-driven approaches for coordination and decision-making. However, coordination among agents, secure task allocation, and scalability in large systems still remain critical challenges. To address these issues, researchers have explored several methods such as leader-follower hierarchies, wherein leader agents define global objectives and delegate subtasks, and middle-agent frameworks that streamline service discovery and coordination among agents~\cite{dorri2018multi}. Generally speaking, all these contributions fall under the umbrella of traditional MAS techniques. Indeed, recent progress in {\em Large Language Models} (LLMs) is now rapidly reshaping the MAS landscape, equipping systems with {\em near-human} reasoning capabilities. When integrated into MASs, LLMs can serve as central reasoning agents, substantially enhancing adaptability, collaboration, and decision-making in dynamic environments. Consequently, such advancements have recently propelled MAS applications into areas such as multimodal reasoning, autonomous GUI navigation, and complex mathematical problem-solving—tasks that were once beyond the scope of traditional MAS approaches~\cite{wang2024survey, zhuge2023mindstorms}.

To capitalize on these breakthroughs, LLM-based MASs typically rely on two core principles: ($i$) a robust, task-specific architecture that maximizes the effectiveness of LLMs, and ($ii$) custom methodologies to embed domain-specific knowledge and adaptive strategies within and among the agents and their surrounding environment. However, prescribing architectural designs {\em a priori} can constrain scalability and limit adaptability across different domains. Moreover, integrating LLMs with external knowledge and tools adds another layer of non-negligible complexity. Lastly, developing an LLM-based MAS from scratch typically presents a steep learning curve, posing significant development and usability challenges, especially for non-experts.

In this paper, we introduce Nexus, a novel open-source Python framework that allows users to easily design MAS architectures in a low-code fashion. In addition, Nexus is lightweight, scalable, and orthogonal to both LLMs and application domains, thereby enabling intelligent automation across a wide variety of tasks. The contributions of our work can be summarized as follows:

\setlist{nolistsep}
\begin{enumerate}[noitemsep]
\item We introduce a flexible multi-supervisor hierarchy for efficient and scalable task delegation among agents. This hierarchy consists of a single root {\em Supervisor} agent that acts as a global orchestrator, along with dedicated {\em Task Supervisors} distributed throughout the MAS structure. This design readily supports a divide-and-conquer approach to solving complex problems while minimizing design effort.
\item We enable support for architecture design and workflow definition using simple, plain-text YAML files, streamlining system design and eliminating the need for extensive programming expertise.
\item We evaluate Nexus-designed architectures on three different domains of application: ($i$) {\bf coding tasks}, where we assess the ability of our MASs to produce correct code in both Python (using the HumanEval dataset~\cite{chen2021evaluating}) and Register-Transfer Level (RTL) design (using the VerilogEval-Human dataset~\cite{liu2023verilogeval}); ($ii$) complex {\bf math problem solving}, by randomly selecting five difficult-level problems from the MATH dataset~\cite{hendrycks2021measuring} to evaluate reasoning and problem-solving capabilities; and ($iii$) {\bf optimization in {\em Electronic Design Automation} (EDA)}, by challenging our framework with a compact architecture capable of autonomously performing timing closure on a set of industry-standard designs from the VTR benchmark suite~\cite{murray2020vtr}.
\item We release the source code of this project on GitHub at \url{https://github.com/PrimisAI/nexus} under a permissive open-source license.
\end{enumerate}

Experimental results demonstrate that Nexus empowers MAS architectures to achieve state-of-the-art performance across multiple domains. In coding tasks, the system attains a 99\% pass rate on HumanEval and 100\% on VerilogEval-Human, thus outperforming recent reasoning language models like o3-mini~\cite{o3mini} and DeepSeek-R1~\cite{guo2025deepseek}, while, at the same time, also excelling in complex reasoning and math problem solving. Moreover, Nexus-based architectures effectively tackle challenging timing closure tasks in EDA, achieving multi-objective optimizations that yield an average power saving of nearly 30\%.

The remainder of the paper is organized as follows: Section~\ref{sec:background} provides a brief introduction to MASs and their evolution from heuristics-based agents to LLM-based agents. Section~\ref{sec:methods} details the internal mechanisms of the proposed Nexus framework, highlighting key differences with respect to existing solutions. Section~\ref{sec:results} illustrates the experimental results, while Section~\ref{sec:conclusions} concludes the paper with a summary of the contributions and findings.

\section{Background \& Related Work}\label{sec:background}
This section provides a concise overview of how MASs have evolved. We begin by summarizing their origins and pinpointing the key milestones that have propelled modern MASs into one of the most promising approaches for achieving advanced integrated intelligence. However, it is worth noting that many recent advances in LLM-based MASs stem from leveraging LLMs as a practical means to interface multiple agents, rather than from a direct extension of heuristics-based MAS methods. In other words, recent developments in the field have largely occurred in parallel with\textemdash or even independently of\textemdash traditional MAS research.

\subsection{The Origin of Multi-Agent Systems}
\begin{figure}[h]
    \centering
    \includegraphics[width=\linewidth]{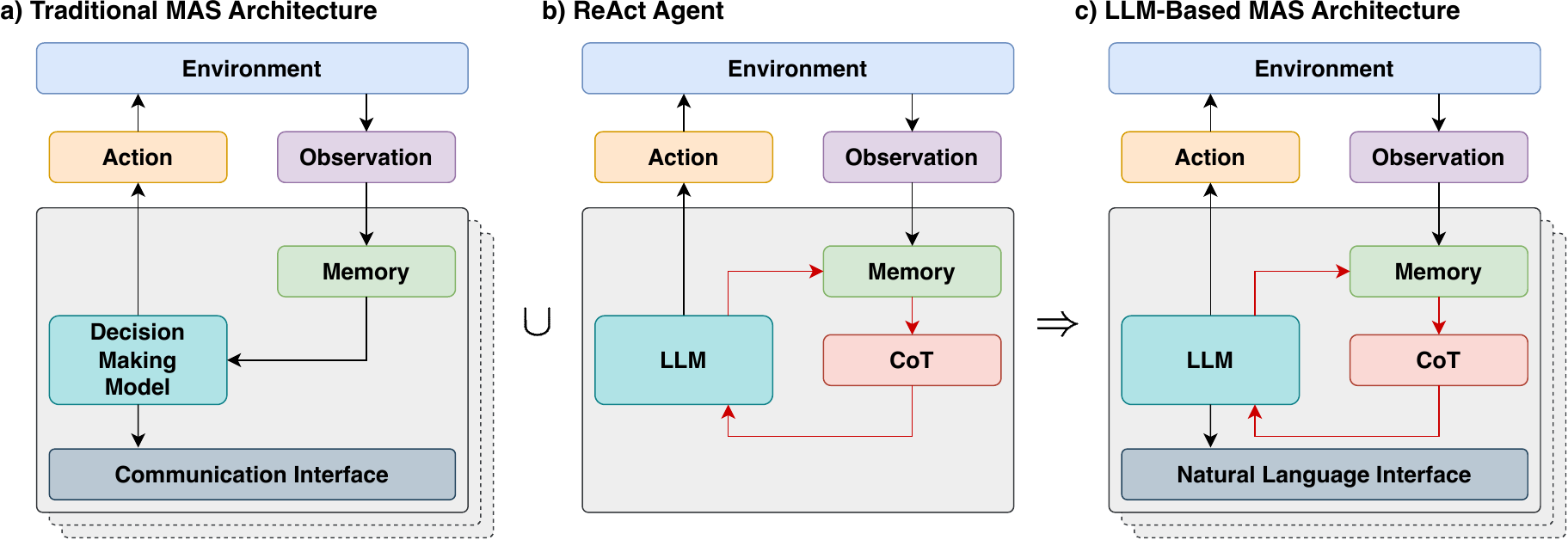}
    \caption{Evolution of Multi-Agent System Architectures: a) {\em Traditional MAS Architecture}, where agents interact with their environment through observations and actions; b) {\em ReAct Architecture}, an innovative agent design that incorporates advanced reasoning capabilities; and c) {\em LLM-Based MAS Architecture}, a cutting-edge approach leveraging LLMs for reasoning and decision-making.}
    \label{fig:mas-architecture}
\end{figure}

As already mentioned earlier, a MAS architecture, depicted in Figure~\ref{fig:mas-architecture}-a, comprises multiple agents capable of perceiving their environment, reasoning about both local states and shared objectives, and executing actions in parallel to achieve a common goal. By distributing tasks among agents, a MAS leverages specialized capabilities and diverse perspectives, often yielding more robust solutions~\cite{durfee1999distributed}. Early conceptualizations of MASs emerged during the study of distributed problem-solving in the 1980s, spurred by the notion that coordinated groups of autonomous entities can achieve more efficient and reliable outcomes than individuals working alone. Foundational work by Minsky~\cite{minsky1988society}, Wooldridge and Jennings~\cite{wooldridge1995intelligent}, and Stone and Veloso~\cite{stone2000multiagent} established the core principles of agent autonomy, collaboration, and decentralized decision-making. Over the ensuing decades, researchers addressed fundamental challenges in agent-to-agent communication, exploring topics such as task allocation~\cite{sandholm1998contract}, negotiation protocols~\cite{jennings2001automated,shoham1997emergence}, and conflict resolution mechanisms~\cite{rosenschein1994rules}.

\subsection{LLM-Enhanced Multi-Agent Systems}
Recent advances in LLMs have reignited interest in MASs by enabling more sophisticated reasoning, natural language communication, and advanced planning. While traditional MASs often relied on symbolic or rule-based methods for coordination and decision-making~\cite{genesereth1994software,genesereth1997agent}, modern LLMs can interpret complex instructions, generate contextually relevant responses, and adapt naturally to diverse communication protocols. This significantly reduces the need for manually crafted dialogue policies and negotiation strategies. Several works have already demonstrated that LLMs can serve as the backbone of MASs. For instance, Park et al.\cite{park2023generative} illustrate how LLM-powered agents simulate dynamic social interactions by reasoning about internal goals and social norms. These agents can generate messages to coordinate with others, interpret feedback, and refine their plans in real time, aligning more seamlessly with human-like communication standards\cite{andreas2022language}. Consequently, they are inherently better equipped to tackle complex tasks requiring contextual understanding, creative reasoning, or dynamic problem-solving.

\subsection{ReAct: Reasoning and Action}
In parallel with the evolution of MASs and LLMs, researchers have sought to render each agent's reasoning process more explicit and adaptable. One notable approach is the {\em ReAct} paradigm~\cite{yao2022react} (short for ``Reasoning + Act''), originally introduced for single-agent systems, as depicted in Figure~\ref{fig:mas-architecture}-b. ReAct structures an agent's decision-making into an iterative cycle that goes as follows: ($i$) {\bf observe}, where the agent receives new information from its environment or other agents; ($ii$) {\bf reason}, where the agent produces a {\em chain-of-thought} (CoT), often internally or in a hidden state, to determine the next step; and ($iii$) {\bf act}, where the agent executes a specific action, such as calling a tool, sending a message, or updating its state. This cycle continues until the task is complete. By explicitly separating reasoning from action, ReAct enhances transparency and adaptability, enabling agents to dynamically revise their approaches as contexts evolve~\cite{wei2022chain}.

\subsection{Next-Generation MAS Architectures}
Although ReAct was originally conceived for single-agent scenarios, its design naturally extends to multi-agent systems, where each agent is supported by an LLM (see Figure~\ref{fig:mas-architecture}-c). In this setting, each agent follows a ReAct-style loop to process observations, perform internal reasoning, and act, such as by communicating with other agents or invoking external tools. This integration yields powerful synergies, including:

\setlist{nolistsep}
\begin{itemize}[noitemsep]
\item {\bf Enhanced Coordination}: LLM-based agents can communicate in natural language to negotiate plans, share partial solutions, or request assistance.
\item {\bf Iterative Reasoning and Action}: The ReAct cycle ensures that each agent's CoT remains flexible, context-aware, and up-to-date as it receives new inputs from the environment or from other agents.
\item {\bf Meta-Cognitive Techniques}: Approaches such as reflection~\cite{yao2022react}, task decomposition~\cite{wei2022chain,yao2024tree}, and dynamic tool creation~\cite{qin2023toolllm} can be layered on top of the ReAct loop to enable deeper analysis, more systematic planning, and specialized behaviors.
\end{itemize}

These developments clearly point towards a future where MASs, enhanced by LLMs and meta-cognitive processes like ReAct, can handle sophisticated teamwork and autonomous problem-solving at scales once deemed intractable for traditional MAS approaches. Indeed, recent work has demonstrated that multi-agent setups are particularly effective for tasks such as GUI automation~\cite{agashe2024agent,tan2024cradle,zhang2024ufo} and automatic code debugging~\cite{sami2024aivril,sami2024eda,zhao2024mage}, just to name a few, illustrating a rapidly evolving landscape of possibilities.

\subsection{Modern MAS Frameworks}
These advancements have led to the emergence of various toolkits and frameworks aimed at simplifying the design and deployment of agentic workflows. Projects such as AutoGPT\footnote{AutoGPT introduced a multi-agent paradigm in its most recent release.}\cite{autogpt} and HuggingGPT\cite{shen2024hugginggpt} offer automated pipelines for task decomposition and tool usage. However, they predominantly rely on single-agent paradigms with modular sub-routine execution rather than on fully decentralized, multi-agent collaboration. Other open-source initiatives, including LangGraph~\cite{langgraph}, AutoGen~\cite{wu2023autogen}, crewAI~\cite{crewai}, Dynamiq~\cite{dynamiq}, Magentic-One~\cite{fourney2024magentic}, and Haystack~\cite{haystack}, provide more customizable infrastructures for building multi-agent systems, although they typically require significant coding expertise. By contrast, commercial offerings sometimes feature no-code interfaces, yet often lack transparent integration paths for broader automation.

In addressing this gap, the proposed Nexus framework offers a twofold solution. First, it streamlines the creation and prototyping of complex agentic structures by means of YAML files, significantly reducing the programming expertise required to design architectures for complex problems. Second, it provides straightforward integration with software automation environments\textemdash typically built on top of shell interfaces\textemdash thereby supporting end-to-end automated workflows out-of-the-box. Designed with maximum flexibility in mind, Nexus accommodates multiple supervisors (or orchestrators) within its hierarchical architecture, facilitating the management of highly complex tasks while enhancing decentralization and collaborative problem-solving. All these features distinguish Nexus from existing frameworks and open new avenues for scalable, automated design and deployment across diverse domains.

\section{Nexus: A Scalable Hierarchical Multi-Agent Architecture}\label{sec:methods}
The Nexus architecture is a modular design that integrates a single root {\em Supervisor} agent alongside multiple {\em Task Supervisors} and {\em Worker} agents. These components are arranged within a hierarchical execution graph to ensure efficient task delegation, flexibility, and scalability. The remainder of this section details the architecture's core components, overall operational workflow, and distinguishing features.

\subsection{Core Components and Structure}
Nexus architectures rely on a single root {\em Supervisor} that mediates interactions between the user and the network of agents. Its primary responsibilities include: ($i$) {\bf task decomposition}, which involves breaking high-level prompts into actionable subtasks; ($ii$) {\bf agent selection}, where subtasks are delegated to the most appropriate {\em Worker} agent (or {\em Task Supervisor}, when instantiated) based on each agent's specialization; and ($iii$) {\bf result aggregation}, i.e., collecting outputs from delegated subtasks and synthesizing them into a cohesive final response.

{\em Worker} agents are specialized problem solvers tasked with executing the subtasks assigned by their supervisor. Each {\em Worker} operates in an isolated environment and possesses a unique {\em specialization} defined by its system message and the associated tools, environment variables, and {\em ad hoc} functions. Their key capabilities include: ($i$) utilizing dedicated tools (e.g., web search, shell commands, file manipulation) or knowledge bases to perform domain-specific operations; ($ii$) iteratively refining intermediate results through interactions with these tools or external data sources; and ($iii$) returning structured, task-specific outputs to the appropriate supervisor upon completion.

In addition to these agents, Nexus incorporates a global {\em Memory} mechanism along with a set of external {\em Tools}. The {\em Memory} works as a centralized repository that stores partial results, metadata, and relevant instructions, ensuring that all agents maintain an up-to-date view of task progress. Although the {\em Memory} serves as a shared repository, Nexus enforces role-based access control: while the {\em Supervisor} has global access, a {\em Worker} agent is confined to its own event history, and a {\em Task Supervisor} can access all memory locations associated with its assigned agents. On the other hand, the external {\em Tools} provide agents with specialized functionalities, such as data processing pipelines or access to external computational resources, thereby fostering a uniform and transparent environment for managing complex workflows.

\begin{figure}[h]
    \centering
    \includegraphics[width=0.6\linewidth]{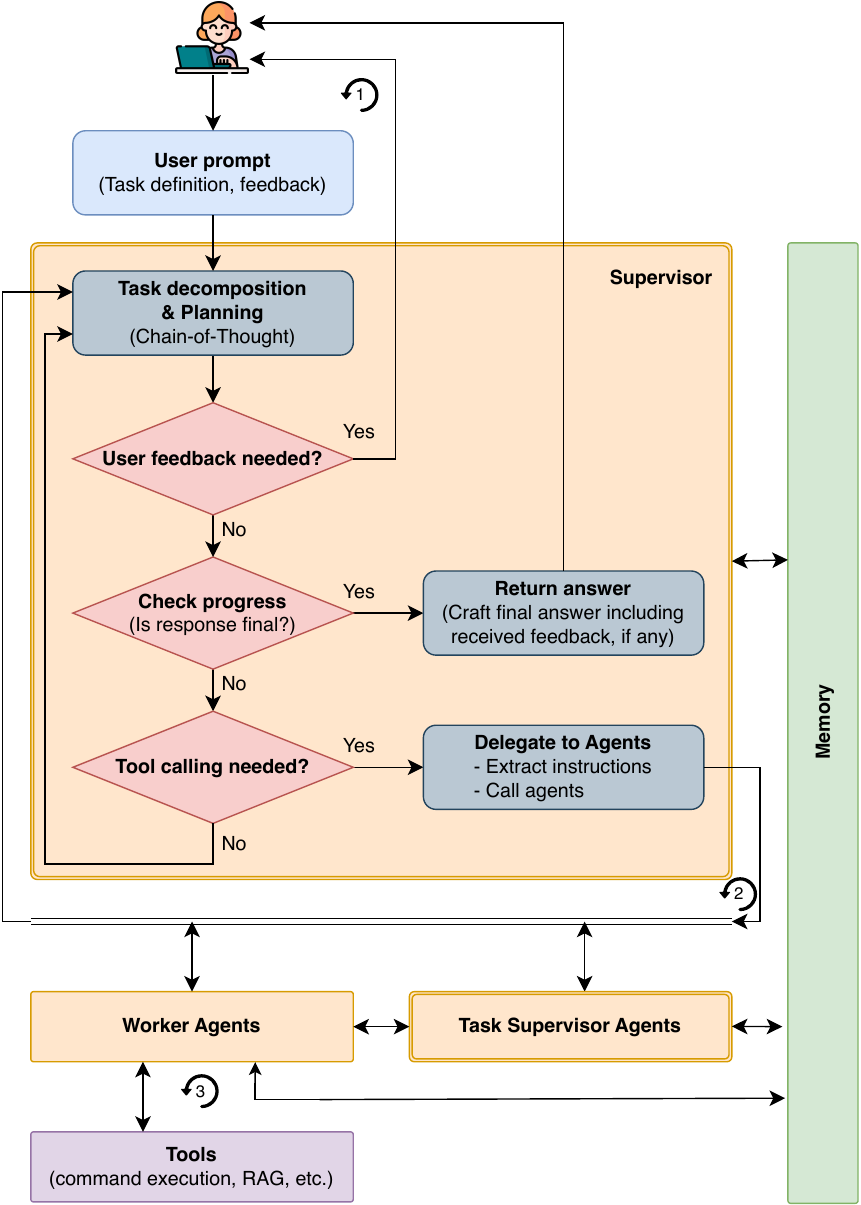}
    \caption{Overview of the Nexus architecture. A root {\em Supervisor} receives user prompts and decides whether to finalize the solution or delegate its execution. Tasks of moderate complexity can be handled by specialized {\em Worker} agents, while particularly intricate tasks can be coordinated by intermediate {\em Task Supervisors}. \emph{Memory} maintains a synchronized record of partial outputs and relevant context. Circled markers denote the three main loops that are entailed in the proposed workflow.}
    \label{fig:nexus}
\end{figure}

Figure~\ref{fig:nexus} depicts the overall architecture and workflow of the proposed Nexus framework. More formally, the architecture is modeled as a rooted directed graph, denoted by 

\begin{equation}
\Gamma = (V, E),
\end{equation}

which captures the relationships between various agents and components within Nexus. The vertex set $V$ is partitioned into the following three disjoint subsets:

\begin{equation}
V = S \cup T \cup W,
\end{equation}

where $S$ represents the set of {\em Supervisor} agents\textemdash with the unique root node $s \in S$\textemdash, $T$ denotes the set of {\em Task Supervisor} agents, and $W$ corresponds to the set of {\em Worker} agents. Based on these assumptions, the edge set

\begin{equation}
E \subseteq V \times V
\end{equation}

captures all relationships among agents defined within any given $\Gamma$. A critical element of this structure is the hierarchical relationship between two or more elements, which can be formalized by introducing a parent function defined as 

\begin{equation}
\varphi \colon V \setminus \{s\} \to V.
\end{equation}

For every node $v \in V \setminus \{p\}$, the directed edge 

\begin{equation}
(\varphi(v), v) \in E
\end{equation}

identifies the immediate supervisor-to-agent relationship between node $v$ and its parent node $\varphi(v)$. This relationship adheres to the following constraints: first, if $v \in T$, i.e., a {\em Task Supervisor}, then its parent $\varphi(v)$ must belong to $S$; second, if $v \in W$, then $\varphi(v)$ is either in $T$ or $S$. Consequently, this design guarantees that every agent, other than the root, has a unique predecessor, thus ensuring that there exists a unique directed path from the root $s$ to any node $v \in V$. In addition to these hierarchical relationships, the architecture also allows the inclusion of extra edges that capture non-hierarchical interactions between components. Although these communication edges do not necessarily conform to the strict parent–child relationship, they are incorporated in such a way as to maintain the overall hierarchical integrity of $\Gamma$. To further clarify the hierarchical structure, let us define a level function as 
\begin{equation}
\ell \colon V \to \mathbb{N},
\end{equation}
which is determined recursively by setting \(\ell(s) = 0\) for the root and, for any other node \(v \in V \setminus \{s\}\), defining 
\begin{equation}
\ell(v) = \ell(\varphi(v)) + 1.
\end{equation}
Such a function assigns each agent a level based on its distance from the root, with the root at level 0, its immediate subordinates at level 1, and so on.

\subsection{Multi-Loop Workflow}
Nexus introduces an iterative process for task decomposition and execution, organized into three primary interaction loops, as depicted in Figure~\ref{fig:nexus} (numbered circled arrow markers).

\underline{\smash{\bf First Loop: User-Supervisor Interaction.}}
In the first loop, the user provides a high-level prompt to the {\em Supervisor}. The {\em Supervisor} interprets the request and outlines an initial task execution plan while soliciting user feedback to ensure that the evolving plan remains aligned with the user's objectives. This iterative exchange continues until the {\em Supervisor} is ready either to delegate subtasks to other agents or to finalize a solution.

\underline{\smash{\bf Second Loop: Supervisor-Agent Coordination.}}
In the second loop, the {\em Supervisor} (or a {\em Task Supervisor}, when applicable) assigns well-defined subtasks to {\em Worker} agents based on their specific characteristics and specialization. The {\em Worker} agents then generate intermediate outputs by interacting with the tools and resources available within their designated working environments. If a {\em Worker} fails to produce satisfactory results or encounters a bottleneck, the {\em Supervisor} revises the subtask instructions or reallocates the task to another agent. This iterative reassessment continues until the subtasks collectively meet the established quality criteria.

\underline{\smash{\bf Third Loop: Intra-Agent Operations.}}
The third loop operates within each {\em Worker} agent's internal environment. Upon receiving its assigned subtask, a {\em Worker} iteratively leverages relevant tools, local data structures, or external knowledge bases to refine its intermediate output. Once a sufficiently polished solution is achieved, the {\em Worker} relays the result back to its supervisor for integration or final synthesis.

In summary, Nexus combines a robust hierarchical framework with the flexibility to support diverse interaction patterns among agents and their operating environments. This design endows the framework with three fundamental properties:

\begin{itemize}
\item \textbf{Scalability\textemdash}The framework can seamlessly incorporate new agents or supervisory nodes as task complexity escalates.
\item \textbf{Modularity\textemdash}{\em Worker} agents operate independently, enabling the straightforward integration or replacement of domain-specific capabilities.
\item \textbf{Robustness\textemdash}Hierarchical delegation and iterative feedback loops minimize the impact of individual agent failures, as tasks can be reassigned or refined with minimal disruption.
\end{itemize}

\subsection{Framework Installation and Basic Usage}

To ensure reproducibility and encourage widespread adoption, Nexus is distributed as an installable Python package. Most users can install Nexus directly from PyPI by executing the following command to retrieve the latest stable release:

\begin{verbatim}
    pip install primisai
\end{verbatim}

Alternatively, developers interested in modifying the framework's core functionalities can clone the GitHub repository and install the package in {\em editable} mode as follows:

\begin{verbatim}
    git clone git@github.com:PrimisAI/nexus.git
    cd nexus
    pip install -e .
\end{verbatim}

Listing~\ref{lst:example_arch} provides an example of how to instantiate a simple three-agent architecture for code refactoring, comprising a {\em Supervisor} and two specialized \emph{Worker} agents.

\begin{lstlisting}[
    language=Python,
    basicstyle=\small\ttfamily,
    numbers=left,
    numberstyle=\tiny\color{gray},
    stepnumber=1,
    numbersep=10pt,
    backgroundcolor=\color{gray!5},
    xleftmargin=18pt,
    framexleftmargin=15pt,
    xrightmargin=5pt,
    frame=single,
    rulecolor=\color{gray},
    breaklines=true,
    breakatwhitespace=true,
    showstringspaces=false,
    tabsize=2,
    caption={A Python implementation demonstrating a three-agent Nexus MAS architecture for automated code refactoring.},
    label={lst:example_arch},
    keywordstyle=\color{blue}\bfseries,
    commentstyle=\color{green!50!black}\itshape,
    stringstyle=\color{orange}]
from primisai.nexus.core import Agent, Supervisor

# Configure large language model parameters.
llm_config = {
    "api_key": "your-api-key-here",
    "model": "model-name-here",
    "base_url": "model-url-here",
}

# Create the root supervisor named "ProgrammingTaskCoordinator".
coordinator = Supervisor("ProgrammingTaskCoordinator", llm_config)

# Instantiate specialized Worker agents with programming-specific system prompts.
code_analyzer = Agent(
    "CodeAnalyzer",
    llm_config,
    system_message="You are a coding expert specialized in static code analysis. Your task is to evaluate code quality, identify potential bugs, and suggest improvements."
)

code_refactorer = Agent(
    "CodeRefactorer",
    llm_config,
    system_message="You are a programming assistant skilled in code refactoring and optimization. Your goal is to enhance code efficiency, readability, and maintainability while preserving functionality."
)

# Register agents with the ProgrammingTaskCoordinator.
coordinator.register_agent(code_analyzer)
coordinator.register_agent(code_refactorer)

# Display the agent hierarchy.
coordinator.display_agent_graph()

# Initiate an interactive session for collaborative programming support.
coordinator.start_interactive_session()
\end{lstlisting}

In order o further highlight Nexus' flexibility, Listing~\ref{lst:example_arch_yaml} presents the same architecture defined via our dedicated YAML file support. This approach not only reduces development complexity and effort, but it also allows users to separate configuration from code, thereby enhancing readability and maintainability.

\begin{lstlisting}[
    language=my-yaml,
    basicstyle=\small\ttfamily,
    numbers=left,
    numberstyle=\tiny\color{gray},
    stepnumber=1,
    numbersep=10pt,
    backgroundcolor=\color{gray!5},
    xleftmargin=18pt,
    framexleftmargin=15pt,
    xrightmargin=5pt,
    frame=single,
    rulecolor=\color{gray},
    breaklines=true,
    breakatwhitespace=true,
    showstringspaces=false,
    tabsize=2,
    caption={A YAML configuration defining a three-agent Nexus MAS architecture for automated code refactoring..},
    label={lst:example_arch_yaml},
    keywordstyle=\color{blue}\bfseries,
    commentstyle=\color{green!50!black}\itshape,
    stringstyle=\color{orange}]
supervisor:
    name: ProgrammingTaskCoordinator
    type: supervisor
    llm_config:
      model: ${LLM_MODEL}
      api_key: ${LLM_API_KEY}
      base_url: ${LLM_BASE_URL}
    system_message: "You are the supervisor for programming tasks. Oversee code analysis and refactoring operations."
    children:
      - name: CodeAnalyzer
        type: agent
        llm_config:
          model: ${LLM_MODEL}
          api_key: ${LLM_API_KEY}
          base_url: ${LLM_BASE_URL}
        system_message: "You are a coding expert specialized in static code analysis. Evaluate code quality, identify bugs, and suggest improvements."
      - name: CodeRefactorer
        type: agent
        llm_config:
          model: ${LLM_MODEL}
          api_key: ${LLM_API_KEY}
          base_url: ${LLM_BASE_URL}
        system_message: "You are a programming assistant skilled in code refactoring and optimization. Enhance code efficiency, readability, and maintainability."
\end{lstlisting}

Interested readers can refer to the GitHub repository, available at \url{https://github.com/PrimisAI/nexus}, for advanced usage, additional instructions, and further examples.

\section{Experimental Results}\label{sec:results}
In this section, we present the experimental evaluation of the proposed Nexus framework. Our objective is to assess both the performance and robustness of the tool across a diverse set of benchmarks and use cases, thereby providing a comprehensive and unbiased analysis of its capabilities. The discussion is organized into three main case studies, each addressing a distinct use case.

\subsection{Methodology}

In all experiments, except those reported in Section~\ref{sec:ppa}, performance was evaluated using the {\em pass rate} (denoted by $\mathcal{A}$), which is defined as the ratio between the number of samples that pass all checks and the total number of samples in the benchmark suite. In some cases, such as with coding-related tasks, we distinguish between $\mathcal{A}_{s}$, i.e., the success rate of solutions passing all syntax checks, and $\mathcal{A}_{f}$, which reflects the success rate of designs that are not only syntactically correct but also functionally accurate. Notably, $\mathcal{A}_{f}$ was determined by executing the tests provided in the benchmark suite, ensuring a comprehensive validation of the proposed approach. For the agents, we employed Claude 3.5 Sonnet v1 or Claude 3.5 Sonnet v2~\cite{claude3.5}. Both models\footnote{Models accessed through AWS Bedrock, with the following identifiers: anthropic.claude-3-5-sonnet-20240620-v1:0 and anthropic.claude-3-5-sonnet-20241022-v2:0.} were configured with a {\em temperature} of 0.7 and a {\em top\_p} of 1.

\subsection{Case Study I: Coding Tasks}

In this section, we assess the effectiveness of the Nexus framework in addressing programming-related tasks. Our evaluation encompasses two benchmark families: HumanEval~\cite{chen2021evaluating}, a suite of 164 problems focused on Python code generation, and VerilogEval-Human~\cite{liu2023verilogeval}, which comprises 156 challenges involving Verilog code generation and verification. Notably, our approach leverages a single, unified Nexus architecture that is consistently applied across both sets of coding challenges.

\begin{figure}[h]
    \centering
    \includegraphics[width=.8\linewidth]{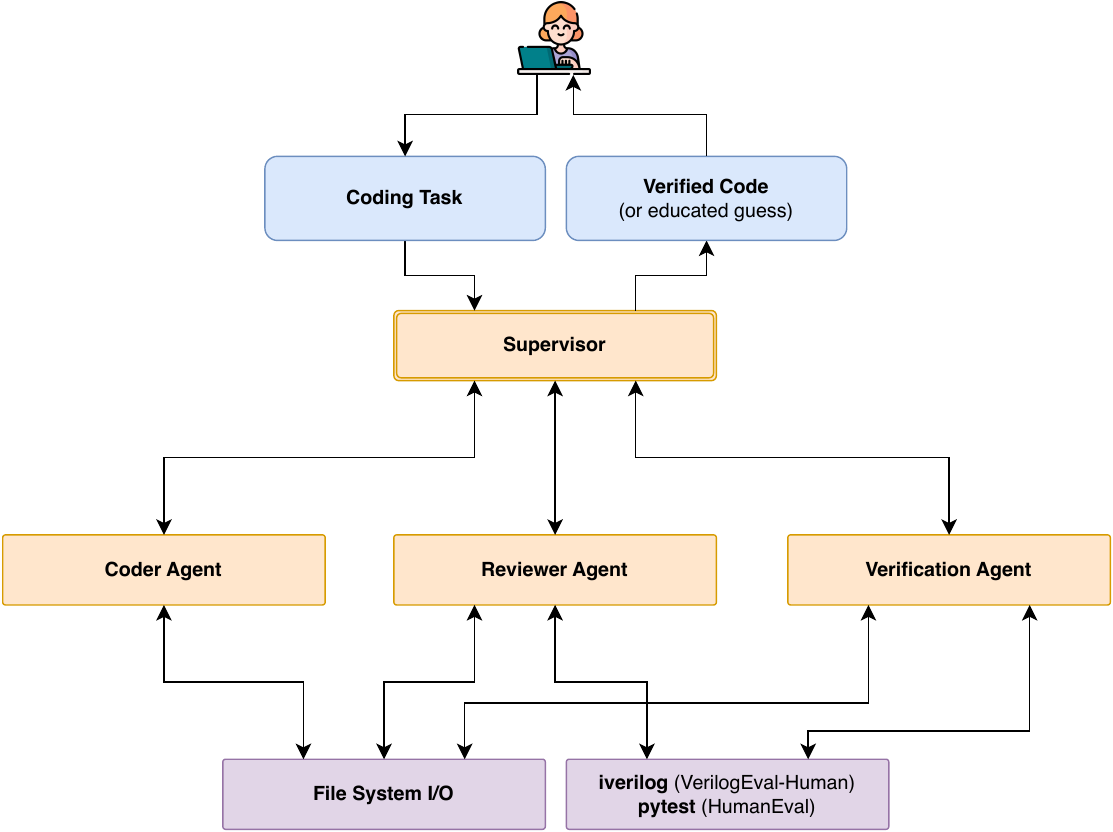}
    \caption{Unified Nexus-based MAS architecture for solving code-related tasks.}
    \label{fig:coding_arch}
\end{figure}

As depicted in Figure~\ref{fig:coding_arch}, the proposed MAS architecture comprises the following core agents: ($i$) a {\em Coder} agent, responsible for generating code solutions (in either Python or Verilog) along with corresponding unit tests or testbenches; ($ii$) a {\em Reviewer} agent, tasked with reviewing and refining code to identify and correct syntax or compilation issues; and ($iii$) a {\em Verification} agent, which executes tests or simulations to assess functionality. As mentioned earlier, the overall structure remains consistent for both benchmark suites, with the specific tools employed (i.e., {\tt pytest} or {\tt iverilog}) chosen to suit the target domain. The workflow proceeds as follows:
\begin{enumerate}
    \item \textbf{Planning \& Task Delegation:} The {\em Supervisor} receives the user's prompt, decomposes the problem into multiple tasks, and assigns the {\em Coder} agent to initiate the first iteration.
    \item \textbf{Code Generation:} The {\em Coder} agent produces the Python (or Verilog) solution along with unit tests (or testbench), storing the output via the \texttt{save\_code} tool, a utility function defined as part of the {\em File System I/O} interface.
    \item \textbf{Syntax Adjustments:} The {\em Reviewer} analyzes the proposed implementation by leveraging the \texttt{get\_code} tool and attempts to execute or compile it using the Python interpreter (or {\tt iverilog}). If syntax errors are detected, corrective prompts are issued and sent back to the {\em Coder}, which iterates on the previous version to generate a revised solution.
    \item \textbf{Functional Adjustments:} Once the loop between the {\em Coder} and {\em Reviewer} concludes and the candidate solution is syntactically correct, the {\em Supervisor} delegates the {\em Verification} agent to run the unit tests using \texttt{pytest} (or again {\tt iverilog}) to assess functionality and correctness. In this stage, any errors are analyzed and communicated back to the {\em Coder} agent for further refinement.
    \item \textbf{Wrap-up:} The {\em Supervisor} collects the verified output and returns the final solution to the user.
\end{enumerate}

It is important to note that this workflow is considered {\em self-verifying}, meaning that the overall system autonomously devises tests without any external input.

\subsubsection{Ablation Study}
Table~\ref{tab:coding_results} summarizes the effectiveness of the proposed self-verifying workflow. For both benchmark suites, we report the Claude 3.5 version used in each experiment (column {\em LLM Version}), along with the syntax pass rate (columns $\mathcal{A}_{s}$) and functional pass rate ($\mathcal{A}_{f}$) achieved by both the baseline Claude 3.5 and by Nexus employing the same underlying LLM.

\begin{table}[!h]
    \centering
    \resizebox{.85\textwidth}{!}{
        \normalsize
        \addtolength\tabcolsep{8pt}
        \begin{tabular}{lrrrrrrr}\toprule
            \multirow{2}{*}{\textbf{Benchmark Suite}} &\multirow{2}{*}{\textbf{LLM Version}} &\multicolumn{2}{c}{\textbf{Baseline}} &\multicolumn{3}{c}{\textbf{Nexus (self-verifying)}} \\\cmidrule(lr{8pt}){3-4} \cmidrule(lr{8pt}){5-7}
            & &\textbf{\textbf{$\mathcal{A}_{s}$}} &\textbf{$\mathcal{A}_{f}$} &\textbf{$\mathcal{A}_{s}$} &\textbf{$\mathcal{A}_{f}$} &\textbf{$\Delta\mathcal{A}_{f}$} \\\midrule
            \textbf{HumanEval} &Claude 3.5 v1 &99.39 &87.80 &100 &96.95 &\textbf{\textcolor{blue}{$\uparrow$ 10.42\%}} \\
            \textbf{HumanEval} &Claude 3.5 v2 &98.78 &92.07 &100 &98.78 &\textbf{\textcolor{blue}{$\uparrow$ 7.28\%}} \\
            \textbf{VerilogEval-Human} &Claude 3.5 v2 &90.38 &67.30 &100 &85.90 &\textbf{\textcolor{blue}{$\uparrow$ 27.63\%}} \\
            \bottomrule
        \end{tabular}
        
    }
    \vspace{8pt}
    \caption{Ablation study results on pass rate, with column {\bf $\Delta\mathcal{A}_{f}$} reporting the percentage improvement of the proposed workflow over the corresponding baseline model in terms of functional pass rate. All values are expressed as percentages.}
    \label{tab:coding_results}
\end{table}

As can be observed, the proposed self-verifying workflow maximizes $\mathcal{A}_{s}$ across all scenarios. Moreover, it substantially enhances functional accuracy in every experiment, achieving a remarkable 27.63\% improvement on the VerilogEval-Human benchmark.

\subsubsection{Comparison with State-of-the-Art Approaches}
Applying LLMs to solve coding challenges is an emerging topic that has already captured significant attention, as these models are rapidly transforming how developers approach programming tasks on a daily basis~\cite{etsenake2024understanding}. Table~\ref{tab:coding_comp} compares the performance of the proposed workflow with the most recent and relevant solutions in this fast-evolving application field.

\begin{table}[!h]
    \centering
    \resizebox{.8\textwidth}{!}{
        \normalsize
        \addtolength\tabcolsep{8pt}
        \begin{tabular}{lrrrrr}\toprule
            \textbf{Benchmark Suite} &\textbf{Technology} &\textbf{Self Verifying} &\textbf{Model} &\textbf{$\mathcal{A}_{f}$} \\\midrule
            \multirow{8}{*}{\textbf{HumanEval}} &L2MAC~\cite{holt2023l2mac} &Yes &GPT-4 &90.2 \\
            &MapCoder~\cite{islam2024mapcoder} &Yes &GPT-4 &93.9 \\
            &AgentCoder~\cite{huang2023agentcoder} &Yes &GPT-4 &96.3 \\
            &LLMDebugger~\cite{zhong2024ldb} &Yes &GPT-4o &98.2 \\
            &LPW~\cite{lei2024planning} &Yes &GPT-4o &98.2 \\
            &QualityFlow~\cite{hu2025qualityflow} &Yes &Claude 3.5 &98.8 \\
            &\cellcolor{blue!25}\textbf{Nexus (this work)} &\cellcolor{blue!25}\textbf{Yes} &\cellcolor{blue!25}\textbf{Claude 3.5} &\cellcolor{blue!25}\textbf{98.8} \\
            &LLMDebugger~\cite{zhong2024ldb} &Yes &o1 &99.4 \\ \midrule\midrule
            \multirow{9}{*}{\textbf{VerilogEval-Human}} &RTLFixer~\cite{tsai2024rtlfixer} &Yes &GPT-3.5 &36.8 \\
            &VeriAssist$^{*}$~\cite{huang2024towards} &Yes &GPT-4 &48.3 \\
            &AIvril~\cite{sami2024aivril} &Yes &Claude 3.5 &67.3 \\
            &AIvril2~\cite{sami2024eda} &Yes &Claude 3.5 &77 \\
            &\cellcolor{blue!25}\textbf{Nexus (this work)} &\cellcolor{blue!25}\textbf{Yes} &\cellcolor{blue!25}\textbf{Claude 3.5} &\cellcolor{blue!25}\textbf{85.9} \\\cmidrule(lr){2-5}
            &VeriAssist$^{*}$~\cite{huang2024towards} &No &GPT-4 &50.5 \\
            &VerilogCoder~\cite{ho2024verilogcoder} &No &GPT-4 &94.2 \\
            &MAGE~\cite{zhao2024mage} &No &Claude 3.5 &94.8 \\
            &\cellcolor{blue!25}\textbf{Nexus (this work)} &\cellcolor{blue!25}\textbf{No} &\cellcolor{blue!25}\textbf{Claude 3.5} &\cellcolor{blue!25}\textbf{100} \\
            \bottomrule
            \end{tabular}
    }
    \vspace{8pt}
    \caption{Comparison of the proposed {\em self-verifying} and {\em non-self-verifying} workflows based on Nexus versus relevant existing solutions. HumanEval numbers have been gathered from the Papers With Code leaderboard~\cite{humaneval_leaderboard}. ($^{*}$) This work employs a dual-mode verification mechanism.}
    \label{tab:coding_comp}
\end{table}

For a fair comparison, we divide the results into three main categories: ($i$) self-verifying MASs designed to solve software programming tasks, as measured by the HumanEval benchmark suite; ($ii$) self-verifying MASs intended to produce and autonomously verify Verilog solutions to help hardware engineers meet their stringent domain requirements; and ($iii$) non-self-verifying MASs that, in addition to the user prompt, incorporate guidance on how to verify the RTL produced by the LLM. For the latter, we introduce a slightly revised workflow compared to the one presented earlier in this section. Instead of generating the testbench autonomously, this approach uses the testbench provided in the benchmark suite as a blueprint to create its own version, following a principle similar to that adopted in previous works~\cite{huang2024towards,ho2024verilogcoder,zhao2024mage}. As the numbers indicate, on HumanEval the proposed workflow ranks {\em ex aequo} in second place behind LLMDebugger~\cite{zhong2024ldb}. This result is particularly noteworthy given that the same workflow, when applied to a completely different programming language, and thus operating sub-optimally relative to its intended domain, remains effective despite being orthogonal to the programming language stack. As a result, when considering the VerilogEval-Human benchmark suite, both the {\em self-verifying} and {\em non-self-verifying} workflows outperform existing solutions. Notably, the {\em non-self-verifying} version achieves a remarkable 100\% accuracy\textemdash a feat that, to the best of the authors' knowledge, has never been achieved before.

\begin{figure}[h]
    \centering
    \includegraphics[width=.5\linewidth]{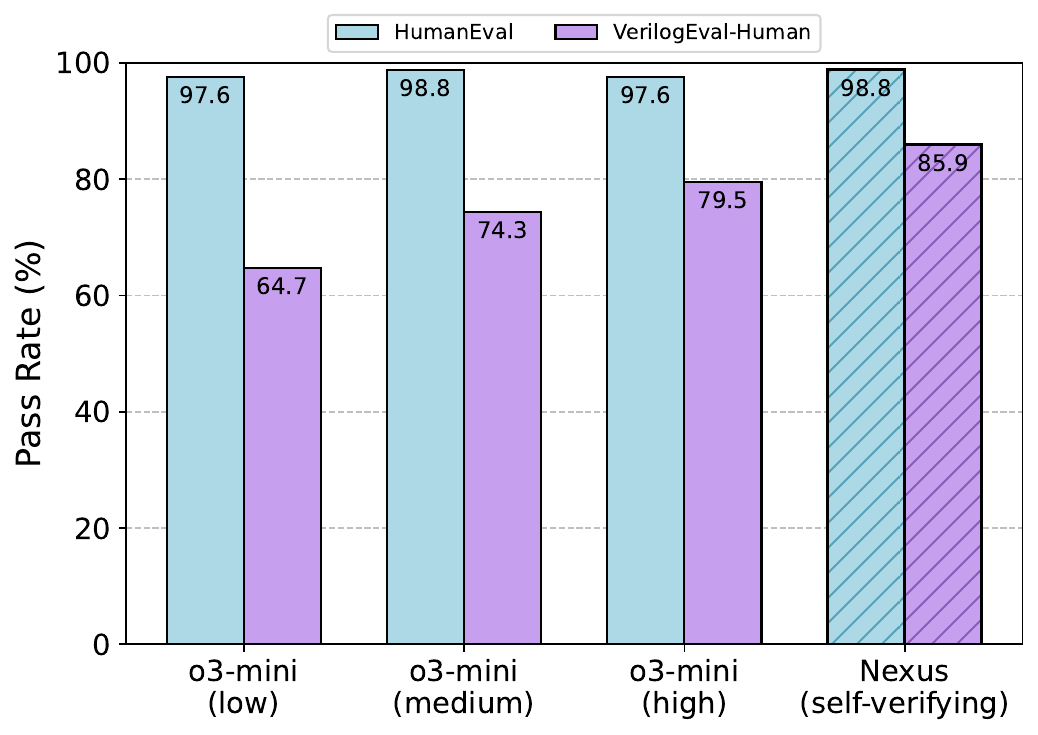}
    \caption{Comparison of reasoning models and the Nexus self-verifying workflow for code-related tasks.}
    \label{fig:reasoning_comp}
\end{figure}

Figure~\ref{fig:reasoning_comp} presents another comparison, this time against emerging {\em reasoning} models\footnote{We originally planned to include DeepSeek-R1~\cite{guo2025deepseek} in our evaluations; however, due to access limitations with DeepSeek's official API, we were only able to obtain results for the VerilogEval-Human dataset, where DeepSeek-R1 achieved a pass rate of 65.38\%.}, i.e., complex systems that leverage self-reflective CoT mechanisms to autonomously decompose tasks into iterative intermediate steps, thereby yielding enhanced accuracy and cost efficiency in solving multi-step problems. In particular, we analyze the recently released o3-mini~\cite{o3mini} from OpenAI\footnote{At the time of writing, o3-mini is the most advanced reasoning model available through the OpenAI API platform.} on both HumanEval and VerilogEval-Human. We consider three levels of reasoning effort\textemdash low, medium, and high\textemdash and compare the resulting pass rates against our proposed self-verifying workflow.

As can be noted, our approach is on par with o3-mini on HumanEval, where all solutions achieve a pass rate exceeding 97\%. In contrast, on VerilogEval-Human, our workflow substantially outperforms all three versions of o3-mini, with the largest margin observed against o3-mini (low) with a 33\% higher pass rate. In conclusion, Nexus enables users to craft workflows that not only match or exceed the performance of state-of-the-art models but also require significantly less effort to create and deploy.

\subsection{Case Study II: Math \& Reasoning Tasks}

To demonstrate the effectiveness of Nexus in solving complex mathematical problems, we conducted a case study using the MATH dataset~\cite{hendrycks2021measuring}. This dataset poses significant challenges for LLMs, particularly due to their limited ability to perform precise calculations without the assistance of external tools.

\begin{figure}[h]
    \centering
    \includegraphics[width=.5\linewidth]{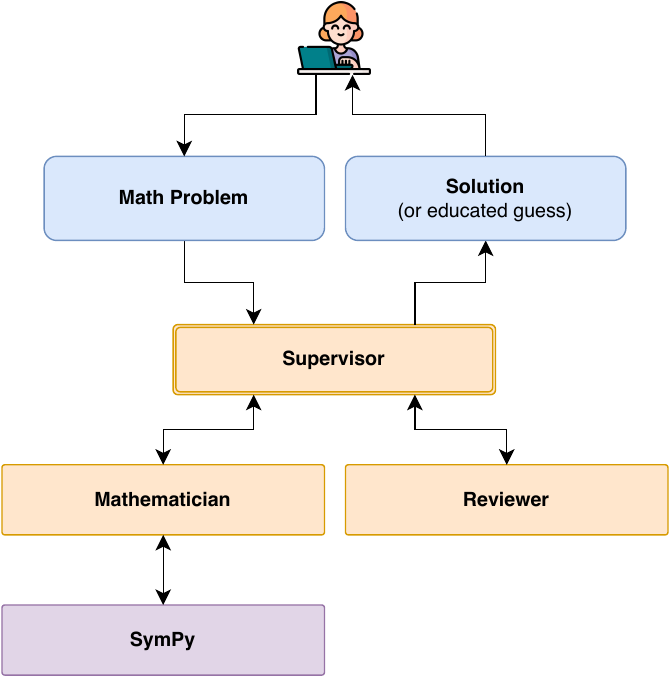}
    \caption{Proposed Nexus-based MAS architecture for solving problems from the MATH dataset.}
    \label{fig:math_arch}
\end{figure}

We devised the Nexus workflow depicted in Figure~\ref{fig:math_arch}, which comprises a {\em Supervisor}, a {\em Mathematician} agent, and a {\em Reviewer} agent, all powered by Claude 3.5 v2. The {\em Supervisor} orchestrates the overall problem-solving process, the {\em Mathematician} generates solutions using the {\tt SymPy} Python package~\cite{meurer2017sympy} for symbolic mathematics, and the {\em Reviewer} evaluates the solutions.

The overall workflow can be detailed as follows:

\begin{enumerate}
\item \textbf{Problem Intake \& Task Assignment:} The {\em Supervisor} receives the math problem from the user, unravels its execution, and assigns tasks to the {\em Mathematician} agent.
\item \textbf{Solution Generation:} The {\em Mathematician} employs {\tt SymPy} to derive a solution.
\item \textbf{Solution Evaluation:} After generating the solution, the {\em Supervisor} forwards both the original problem and the proposed solution to the {\em Reviewer}.
\item \textbf{Feedback and Correction:} The {\em Reviewer} assesses the solution and provides detailed feedback. The {\em Supervisor} then uses this feedback to request corrections if necessary, thereby iterating the process until the solution meets the desired accuracy.
\end{enumerate}

\begin{table}[!h]
    \centering
    \resizebox{.9\textwidth}{!}{
        \normalsize
        \addtolength\tabcolsep{8pt}
        \begin{tabular}{lrrr}\toprule
            \textbf{Problem \#ID} & \textbf{Baseline} & \textbf{Nexus} & \textbf{Observations} \\\midrule
            \textbf{Number Theory \#227} & Failed~\testfailed & Passed~\testpassed & {\tt SymPy} tool crucial \\
            \textbf{Algebra \#2} & Failed~\testfailed & Passed~\testpassed & Baseline far from accurate \\
            \textbf{Geometry \#1140} & Passed~\testpassed & Passed~\testpassed & - \\
            \textbf{Intermediate Algebra \#24256} & Passed~\testpassed & Passed~\testpassed & - \\
            \textbf{Counting and Probability \#25780} & Failed~\testfailed & Passed~\testpassed & Reviewer identified logical flaws \\
            \bottomrule
            \end{tabular}
    }
    \vspace{8pt}
    \caption{Ablation study on the MATH dataset. Columns {\em Baseline} and {\em Nexus} report the assessment of the final answer, comparing it to the ground truth response for both Claude 3.5 v2 and our proposed architecture, respectively. Column {\em Observations} highlights the most significant aspect that emerged from each experiment.}
    \label{tab:math_results}
\end{table}

As summarized in Table~\ref{tab:math_results}, we evaluated this workflow on five randomly selected level-5 problems (i.e., the most challenging problems) drawn from various categories of the MATH dataset. The annotated results indicate whether each system's final output passed the validity check and was deemed correct. The results demonstrate that the proposed Nexus-based workflow successfully solved all five problems, whereas the baseline LLM correctly solved only two.

To better illustrate the effectiveness of the employed workflow in greater detail, let us focus on problem \#25780 from the Counting and Probability category. This problem presents a complex scenario involving probabilistic elevator movements in a 15-story hotel, and its complete definition is provided below:

\begingroup
\renewcommand{\theproblem}{\#25780}
\begin{problem}
    While staying in a 15-story hotel, Polya plays the following game. She enters an elevator on the 6th floor. She flips a fair coin five times to determine her next five stops. Each time she flips heads, she goes up one floor. Each time she flips tails, she goes down one floor. What is the probability that each of her next five stops is on the 7th floor or higher? Express your answer as a common fraction.
\end{problem}
\endgroup

This problem challenges the solver to account for several key aspects: the starting position (6th floor), the probabilistic nature of movements determined by coin flips, the constraint of remaining on the 7th floor or higher for all five stops, and the need to compute the probability of all valid sequences. In essence, the complexity arises from the interplay between the random walks induced by coin flips and the geometric constraints imposed by the floor numbers, thus resulting in a particularly challenging task for typical LLMs.

In this scenario, the baseline LLM erroneously concluded that all five moves must be upward, resulting in a probability of $(\frac{1}{2})^5 = \frac{1}{32}$. The model's reasoning is summarized as follows:

\begin{thinking}[CoT Excerpt\textemdash Baseline (Claude 3.5 v2).]
Assuming each heads (H) moves up one floor and each tails (T) moves down one floor, to ensure we're always at floor 7 or higher:
\begin{itemize}
    \item First flip must be H
    \item Second flip must be H
    \item Third flip must be H
    \item Fourth flip must be H
    \item Fifth flip must be H
\end{itemize}
We need all heads for this to work. Any tails at any point would drop us below floor 7 at some step.
\end{thinking}

This demonstrates the limitations of LLMs in complex probability calculations without external aids. The baseline model failed to consider valid scenarios where some downward movements are possible while still maintaining the condition of staying on the 7th floor or higher. In contrast, the proposed Nexus-based workflow, while initially struggling, eventually produced the correct answer. What follows is the last review phase submitted by the {\em Reviewer} agent to the {\em Supervisor}:

\begin{thinking}[CoT Excerpt\textemdash Nexus (Reviewer)]
After carefully reviewing this final solution, here is my detailed analysis:

\begin{enumerate}[label=\arabic*.]
    \item \textbf{Systematic Sequence Analysis:}
    \begin{itemize}[label={[\texttt{OK}]}]
        \item Correctly starts with H (required to reach 7th floor)
        \item Systematically examines all possible sequences starting with H
        \item Properly tracks floor numbers at each stop
        \item Good organization showing each sequence's path
    \end{itemize}
    
    \item \textbf{Validity Check of Each Sequence:}
    \begin{itemize}[label={[\texttt{OK}]}]
        \item HHHHH: 7,8,9,10,11 (all stops $\geq 7$)
        \item HHHHT: 7,8,9,10,9 (all stops $\geq 7$)
        \item HHHTH: 7,8,9,8,9 (all stops $\geq 7$)
        \item HHHTT: 7,8,9,8,7 (all stops $\geq 7$)
        \item HHTHH: 7,8,7,8,9 (all stops $\geq 7$)
        \item HHTHT: 7,8,7,8,7 (all stops $\geq 7$)
        \item Correctly identifies invalid sequences (HHTTH, HHTTT, HTHXX, HTTXX)
    \end{itemize}
    
    \item \textbf{Completeness Check:}
    \begin{itemize}[label={[\texttt{OK}]}]
        \item All possible sequences starting with H are accounted for
        \item Correctly eliminates sequences going below 7th floor
        \item Properly branches through all possibilities
        \item No valid sequences appear to be missed
    \end{itemize}
    
    \item \textbf{Verification Process:}
    \begin{itemize}[label={[\texttt{OK}]}]
        \item Shows floor numbers at each stop
        \item Clear marking of valid (\texttt{[OK]}) and invalid (\texttt{[X]}) sequences
        \item Explains why invalid sequences are rejected
        \item Demonstrates that all 6 valid sequences stay $\geq 7$th floor
    \end{itemize}
    
    \item \textbf{Probability Calculation:}
    \begin{itemize}[label={[\texttt{OK}]}]
        \item Correct denominator: $2^5 = 32$ (all possible 5-move sequences)
        \item Correct numerator: 6 (verified valid sequences)
        \item Correct reduction: $\frac{6}{32} = \frac{3}{16}$
    \end{itemize}
\end{enumerate}

\bigskip

\textbf{Overall Assessment:}\\[0.5em]
This solution is mathematically correct and complete. It:
\begin{itemize}
    \item Uses a systematic approach
    \item Correctly identifies all valid sequences
    \item Properly verifies the $\geq 7$th floor requirement
    \item Accurately calculates the final probability
    \item Provides clear reasoning and documentation
\end{itemize}

The final answer of $\frac{3}{16}$ is correct and properly justified.
\end{thinking}

As can be noted, the {\em Reviewer} verified that each valid sequence maintains the required height condition and confirmed the soundness of the reasoning and methodology. As a result, the {\em Mathematician}, aided by {\tt SymPy}, correctly determined the probability as 6 (valid sequences) out of 32 (total possible sequences), resulting in $\frac{6}{32} = \frac{3}{16}$.

The complete problem-solving process, including intermediate steps and agent interactions, is presented in Appendix~\ref{appendix:math-problem}, while the complete architecture definition is reported in Appendix~\ref{appendix:yaml-math}. The results demonstrate that Nexus can significantly outperform standalone Claude 3.5 Sonnet in solving complex mathematical problems, showcasing its potential for enhancing AI-driven problem-solving in various domains. By leveraging the strengths of the underlying LLM within a structured multi-agent framework and incorporating external tools, Nexus achieves a level of mathematical problem-solving capability that surpasses the baseline model's performance.

\subsection{Case Study III: Automated Timing Closure in EDA Applications}\label{sec:ppa}
Achieving timing closure and optimizing resource utilization are fundamental challenges in hardware design. In practice, synthesis, placement, and routing strategies are routinely employed to balance stringent timing constraints with efficient hardware resource usage, a balance that is critical for the successful deployment of complex applications on modern computer architectures. To assess the efficacy of our Nexus framework, we conducted extensive experiments using benchmark designs from the well-established VTR benchmark suite~\cite{murray2020vtr}. These benchmarks span a diverse range of application domains, including computer vision ({\tt stereovision0}, {\tt stereovision1}), signal processing ({\tt diffeq1}, {\tt diffeq2}), cryptography ({\tt sha}), and various encoding-decoding applications. In our experimental setup, we leveraged the Xilinx Vivado 2020 Design Suite (hereafter, Vivado) for synthesis, placement, and routing, with all implementations targeting the Xilinx Alveo U200 card.

\begin{figure}[h]
    \centering
    \includegraphics[width=.5\linewidth]{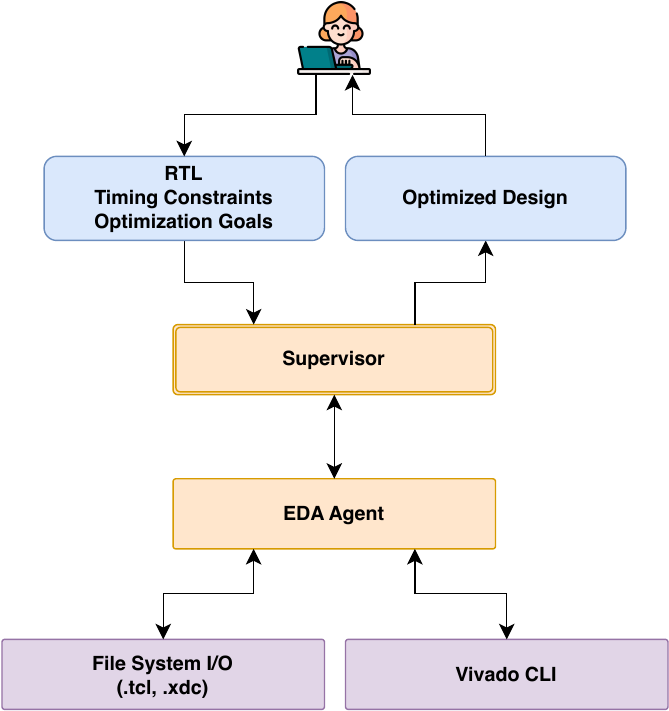}
    \caption{Proposed Nexus architecture for achieving timing closure and design optimization in EDA.}
    \label{fig:synth-tasks}
\end{figure}

For the baseline, the designs were synthesized and mapped by specifying only its target timing constraints, without employing additional optimization techniques. On the other hand, Figure~\ref{fig:synth-tasks} illustrates the adopted Nexus-based architecture, in which all agents are powered by Claude 3.5 v2. The workflow is summarized as follows:
\begin{itemize}
\item \textbf{User Input:} The process begins with the user providing the RTL code along with a prompt that details the timing constraints and optimization goals.
\item \textbf{Analysis:} The \emph{Supervisor} processes these inputs to generate design constraints and commands specifically tailored to meet the timing requirements using Vivado.
\item \textbf{Execution:} The generated constraints and commands are forwarded to the \emph{EDA} agent, which writes them to tool-specific files (e.g., {\tt .xcd} for constraints and {\tt .tcl} for commands) and interfaces directly with the Vivado \emph{Command-Line Interface} (CLI). The \emph{EDA} agent then issues these commands and retrieves reports that provide detailed information on resource utilization, power consumption, timing metrics, and critical paths.
\item \textbf{Feedback:} The \emph{Supervisor} reviews the reports and iteratively refines the optimization strategy until timing closure is achieved or no further improvements are possible.
\end{itemize}

Table~\ref{tab:synth_results} reports the figures of merit for each benchmark. In particular, the column {\em Frequency} reports the target frequency for each benchmark, columns {\em LUTs} and {\em FFs} report resource utilization, while column {\em WSN} reports the worst negative slack and the last column reports the total power. As can be noted, across all benchmarks, the proposed Nexus-based architecture not only achieved timing closure at the target frequencies, but it also delivered significant improvements in key metrics, with an average LUT reduction of 26.64\% and a power reduction of nearly 30\%.

\begin{table}[!h]
    \centering
    \resizebox{\textwidth}{!}{
        \huge
        \addtolength\tabcolsep{8pt}
        \begin{tabular}{lrrrrrrrrrr}\toprule
            \multirow{2}{*}{\textbf{Design}} &\multirow{2}{*}{\textbf{Frequency (MHz)}} &\multicolumn{2}{c}{\textbf{LUTs}} &\multicolumn{2}{c}{\textbf{FFs}} &\multicolumn{2}{c}{\textbf{WNS (ns)}} &\multicolumn{2}{c}{\textbf{Power (W)}} \\\cmidrule(lr{8pt}){3-4} \cmidrule(lr{8pt}){5-6} \cmidrule(lr{8pt}){7-8} \cmidrule(lr{8pt}){9-10}
            & &\textbf{Baseline} &\textbf{Nexus} &\textbf{Baseline} &\textbf{Nexus} &\textbf{Baseline} &\textbf{Nexus} &\textbf{Baseline} &\textbf{Nexus} \\\midrule
            \textbf{diffeq1} &150 &357 &345 &209 &209 &-0.187~\testfailed &0.022~\testpassed &2.634 &2.6 \\
            \textbf{blob\_merge} &200 &5400 &5227 &575 &575 &0.402~\testpassed &0.0384~\testpassed &2.512 &2.51 \\
            \textbf{stereovision0} &333 &3959 &3176 &10290 &7595 &0.5~\testpassed &0.313~\testpassed &2.995 &3 \\
            \textbf{stereovision1} &200 &13321 &1281 &11843 &6186 &1.269~\testpassed &1.42~\testpassed &2.963 &2.9 \\
            \textbf{diffeq2} &167 &229 &232 &111 &111 &0.011~\testpassed &0.032~\testpassed &2.618 &2.621 \\
            \textbf{sha} &300 &1031 &998 &895 &895 &0.3~\testpassed &0.298~\testpassed &2.577 &2.551 \\
            \textbf{stereovision2} &154 &9862 &8062 &13589 &17619 &0.403~\testpassed &0.8~\testpassed &3.268 &3.14 \\
            \textbf{stereovision3} &500 &57 &78 &99 &144 &0.9~\testpassed &0.806~\testpassed &2.493 &0.61 \\
            \textbf{mkPktMerge} &500 &12 &16 &16 &16 &0.389~\testpassed &0.174~\testpassed &3.904 &2.01 \\
            \textbf{mkSMAdapter4B} &200 &910 &885 &859 &865 &1.363~\testpassed &0.92~\testpassed &2.612 &2.62 \\
            \textbf{LU8PEEng} &65 &14581 &13740 &5703 &3569 &-11.409~\testfailed &0.24~\testpassed &2.905 &0.98 \\
            \textbf{bgm} &200 &10801 &10292 &5063 &6150 &0.257~\testpassed &0.133~\testpassed &2.947 &0.98 \\
            \textbf{boundtop} &770 &221 &218 &205 &444 &0.459~\testpassed &0.279~\testpassed &2.5 &0.685 \\
            \textbf{ch\_intrinsics} &1250 &25 &25 &90 &122 &-0.029~\testfailed &0.018~\testpassed &2.68 &0.768 \\\midrule
            \textcolor{blue}{\textbf{$\Delta$Avg.}} & & &\textcolor{blue}{\textbf{-26.64\%}} & &\textcolor{blue}{\textbf{-10.19\%}} & & & &\textcolor{blue}{\textbf{-29.37\%}} \\
            \bottomrule
            \end{tabular}
    }
    \vspace{8pt}
    \caption{Results for timing closure tasks. Symbols in column {\tt WSN} indicate whether timing constraints were met (\testpassed) or not (\testfailed).}
    \label{tab:synth_results}
\end{table}

Among the benchmarks, the {\tt LU8PEEng} case study stands out as a compelling demonstration of our proposed architecture's optimization capabilities. This design posed significant challenges for the baseline strategy, with a WNS of -11.409ns observed at 65MHz. In contrast, our architecture achieved a positive slack of 0.24ns while maintaining the target frequency. Additionally, the framework improved resource utilization: LUT usage decreased by 5.77\% (from 14,581 to 13,740) and flip-flop utilization dropped by 37.42 (from 5,703 to 3,569). This dual optimization of timing and resources highlights the framework's ability to effectively navigate complex design trade-offs. Furthermore, the architecture strategically reallocated block RAM resources, increasing BRAM utilization from 42 to 71 units while maintaining DSP usage at 16 units, to achieve optimal implementation. Notably, these improvements were accompanied by a substantial 3$\times$ reduction in power consumption, lowering it from 2.9W to 0.98W. This significant power optimization, alongside enhanced timing and maintained functionality, underscores the architecture's ability to efficiently exploit advanced features in professional-grade EDA tools, thereby enabling the simultaneous achievement of multiple competing optimization objectives without human intervention. Detailed step-by-step interactions observed in this experiment are provided in Appendix~\ref{appendix:ppa-chat}.

\section{Conclusions}\label{sec:conclusions}
In this paper, we introduced Nexus, a lightweight Python framework designed to simplify the construction and management of LLM-based MASs. Nexus addresses critical challenges in MAS development by enabling flexible multi-supervisor hierarchies, offering a streamlined YAML-based architecture and workflow definition, and facilitating seamless integration and extensibility through its modular, open-source design.

Our experimental evaluations demonstrated that Nexus-driven architectures achieve state-of-the-art performance across a range of tasks. In coding domains, our systems attained a 99\% pass rate on HumanEval and 100\% on VerilogEval-Human, outperforming contemporary reasoning language models like o3-mini and DeepSeek-R1. Additionally, Nexus-based systems showcased strong capabilities in complex reasoning and mathematical problem solving. In the context of EDA, our approach effectively managed challenging timing closure tasks while delivering multi-objective optimizations that resulted in average power savings of nearly 30\%.

In conclusion, Nexus represents a significant advancement in the development of MASs, opening new avenues for leveraging large language models in solving complex, real-world challenges.

\bibliographystyle{IEEEtran}
\bibliography{refs}
\end{spacing}

\clearpage
\begin{appendices}
\section{Detailed Problem-Solving Process for MATH Dataset}\label{appendix:math-problem}

\subsection{Problem Statement}
\vspace{-0.5em}
\begin{userbox}
\begin{lstlisting}
User: While staying in a 15-story hotel, Polya plays the following game. She enters an elevator on the 6th floor. She flips a fair coin five times to determine her next five stops. Each time she flips heads, she goes up one floor. Each time she flips tails, she goes down one floor. What is the probability that each of her next five stops is on the 7th floor or higher? Express your answer as a common fraction.
\end{lstlisting}
\end{userbox}

\subsection{Initial Discussion}
\vspace{-0.5em}
\begin{supervisorbox}
\begin{lstlisting}
Supervisor thinking: We need to determine the probability that after starting from the 6th floor, all five stops will be on the 7th floor or higher after a sequence of coin flips. This requires careful consideration of all possible paths and valid outcomes.

Supervisor to the mathematician:
Instruction: Please solve this probability problem with the following steps:
1. Clarify the initial conditions and requirements
2. Determine what needs to happen for all stops to be on 7th floor or higher
3. Calculate the probability considering all valid sequences
4. Express the final answer as a common fraction

Remember that:
- Starting point is 6th floor
- Each H moves up one floor, each T moves down one floor
- Need 5 stops, each must be >= 7th floor
- Coin is fair (P(H) = P(T) = 1/2)
- Building has 15 floors (upper bound)

\end{lstlisting}
\end{supervisorbox}

\subsection{Mathematician's Initial Solution}
\vspace{-0.5em}
\begin{mathematicianbox}
\begin{lstlisting}
Mathematician: Let me solve this probability problem step by step:

1. Initial Conditions Clarification:
   - Start at 6th floor
   - Need exactly 5 stops
   - Each stop must be at 7th floor or higher
   - Each H moves up (+1), each T moves down (-1)
   - Fair coin: P(H) = P(T) = 1/2
   - Building has 15 floors (won't affect our calculation as we won't reach it)

2. Analysis for Stops >= 7th Floor:
   - First stop: Need to be at 7th floor or higher, so must start with H
   - Second stop: Must maintain position >= 7th floor
   - Third stop: Must maintain position >= 7th floor
   - Fourth stop: Must maintain position >= 7th floor
   - Fifth stop: Must maintain position >= 7th floor

3. Probability Calculation:
   - Total possible sequences length 5: 2^5 = 32 (denominator)
   - For numerator, we need to count favorable sequences:
   - For first position: Must be at 7 (floor 6 + 1H)
   - For second position: Must be at least 7 (can be 7,8)
   - For third position: Must be at least 7 (can be 7,8,9)
   - For fourth position: Must be at least 7 (can be 7,8,9,10)
   - For fifth position: Must be at least 7 (can be 7,8,9,10,11)
   - Multiply these possibilities: 1 * 2 * 3 * 4 * 5 = 120
   - Total probability = 3465/32

4. Final Answer:
   The probability is 3465/32

This can be interpreted as roughly 108.28125, meaning there are approximately 108 favorable outcomes for every single unfavorable outcome. This high number makes sense because once we get to the 7th floor or higher, there are many ways to stay above the 7th floor with various combinations of H and T.

Note: The calculation (6 + 1)*(6 + 2)*(6 + 3)*(6 + 4)*(6 + 5)/32 represents the number of possible favorable paths divided by the total number of possible paths (2^5 = 32). Each term represents the number of possible positions at each stop that satisfy our condition of being at or above the 7th floor.
\end{lstlisting}
\end{mathematicianbox}

\subsection{Review Request}
\vspace{-0.5em}
\begin{supervisorbox}
\begin{lstlisting}
Supervisor thinking: I need to review the Mathematician's solution for accuracy and clarity in solving this probability problem involving an elevator game with coin flips.

Supervisor to the reviewer:
Instruction: Please review the Mathematician's solution with particular attention to:
1. The correctness of the probability calculation
2. The logic of requiring all stops to be >= 7th floor
3. The counting method used
4. The final answer and its reasonableness
5. Any potential errors or oversights in the reasoning

\end{lstlisting}
\end{supervisorbox}

\subsection{Reviewer's Analysis}
\vspace{-0.5em}
\begin{reviewerbox}
\begin{lstlisting}
Reviewer: After reviewing the provided solution, I've identified several concerns:

1. Major Calculation Error:
- The solution concludes with 3465/32, but earlier calculates 120 favorable outcomes
- There's no explanation of how 3465 was derived from the previous steps
- These numbers are inconsistent with each other

2. Logical Issues:
- While correctly noting that the first move must be H (to reach 7th floor), the solution doesn't properly account for maintaining position >= 7th floor
- The multiplication 1 * 2 * 3 * 4 * 5 = 120 seems to be counting something, but it's not clear what these numbers represent
- The reasoning for why there are 2, 3, 4, and 5 possibilities at each subsequent step isn't explained

3. Methodology Concerns:
- The solution doesn't track the actual floor numbers reached after each move
- It doesn't consider that after reaching the 7th floor, some combinations of H and T would take us below the 7th floor (which aren't allowed)
- The counting method appears to be oversimplified and doesn't properly account for the path constraints

4. Missing Elements:
- No verification that paths stay at or above 7th floor throughout
- No explanation of how the final number 3465 was obtained
- No sample paths to demonstrate the counting method

5. Recommendations:
- The solution needs to:
  * Track actual floor numbers after each move
  * Consider all valid sequences that stay >= 7th floor
  * Show detailed counting of valid sequences
  * Provide clear explanation for each step
  * Include verification of the final answer

Based on these issues, I cannot confirm that 3465/32 is the correct answer. A complete recalculation with proper path tracking and constraints would be needed for the correct solution.

\end{lstlisting}
\end{reviewerbox}

\subsection{Revised Solution Request}
\vspace{-0.5em}
\begin{supervisorbox}
\begin{lstlisting}
Supervisor thinking: The Reviewer has identified several issues with the previous solution. We need a more rigorous approach that carefully tracks floor numbers and valid sequences.

Supervisor to the mathematician:
Instruction: Please provide a revised solution that:
1. Carefully tracks the floor numbers after each move
2. Lists all valid sequences that keep all stops at or above the 7th floor
3. Shows clear counting of valid sequences
4. Provides detailed explanation for each step
5. Verifies the final answer

Remember:
- Start at 6th floor
- Each H moves up one floor, each T moves down one floor
- Need all 5 stops to be at 7th floor or higher
- P(H) = P(T) = 1/2

\end{lstlisting}
\end{supervisorbox}

\subsection{Mathematician's Revised Solution}
\vspace{-0.5em}
\begin{mathematicianbox}
\begin{lstlisting}
Mathematician: Let me provide a detailed, step-by-step solution:

1. Initial Analysis:
   - Start: 6th floor
   - Need: All 5 stops >= 7th floor
   - Moves: H (+1), T (-1)

2. First Move Requirements:
   - Must start with H to reach 7th floor (6+1=7)
   - Starting with T would go to 5th floor, making it impossible to have all stops >= 7

3. Tracking All Possible Sequences:
   Starting at 6, let's track each sequence:
   H (7) must be first. Then from 7:
   - HH -> 8
   - HT -> 6 (invalid)
   - TH -> 7
   - TT -> 5 (invalid)

   From these positions, let's continue valid paths:
   After HH (8):
   - HHH -> 9
   - HHT -> 7
   - HTH -> invalid (came from 6)
   - HTT -> invalid

   After TH (7):
   - THH -> 8
   - THT -> 6 (invalid)
   - TTH -> invalid
   - TTT -> invalid

4. Complete Valid Sequences (showing floor numbers):
   HHHHH: 6->7->8->9->10->11
   HHHHT: 6->7->8->9->10->9
   HHHTH: 6->7->8->9->8->9
   HHTHH: 6->7->8->7->8->9
   HTHTH: 6->7->6->7->6->7

   Total valid sequences: 5

5. Probability Calculation:
   - Total possible sequences of length 5: 2^5 = 32
   - Valid sequences: 5
   - Probability = 5/32

Verification:
1. Each valid sequence starts with H (required to reach 7th floor)
2. Every stop in valid sequences is >= 7th floor
3. No other sequences are possible while maintaining all stops >= 7
4. Each sequence has exactly 5 moves
5. The probability makes sense as it's less than 1/4 (which would be 8/32)

Final Answer: The probability is 5/32 (approximately 0.15625 or about 15.625%)

This is significantly different from my previous incorrect answer. The error in the previous solution was in assuming that the number of possibilities multiplied at each step. In reality, we need to track each specific sequence and verify that EVERY stop along the way stays at or above the 7th floor, not just the final position.

The key insight is that once you go below 7th floor at any stop, that sequence becomes invalid, regardless of whether you could get back above 7 later. This strict requirement dramatically reduces the number of valid sequences to just 5 out of the possible 32 sequences.
\end{lstlisting}
\end{mathematicianbox}

\subsection{Review Request}
\vspace{-0.5em}
\begin{supervisorbox}
\begin{lstlisting}
Supervisor thinking: I need to review the Mathematician's revised solution which gives a probability of 5/32.

Supervisor to the reviewer:
Instruction: Please review the revised solution, checking:
1. The completeness of sequence tracking
2. The validity of each counted sequence
3. Whether any valid sequences were missed
4. The correctness of the probability calculation
5. The overall logic and reasoning

\end{lstlisting}
\end{supervisorbox}

\subsection{Reviewer's Analysis}
\vspace{-0.5em}
\begin{reviewerbox}
\begin{lstlisting}
Reviewer: After carefully reviewing the revised solution, here is my detailed analysis:

1. Sequence Tracking Method:
[OK] The solution correctly starts at 6th floor
[OK] Properly tracks floor numbers after each move
[OK] Systematically follows branches of the decision tree
[OK] Good elimination of invalid paths that go below 7th floor

2. Validity of Counted Sequences:
[OK] HHHHH: 6->7->8->9->10->11 (valid)
[OK] HHHHT: 6->7->8->9->10->9 (valid)
[OK] HHHTH: 6->7->8->9->8->9 (valid)
[OK] HHTHH: 6->7->8->7->8->9 (valid)
[X]  HTHTH: 6->7->6->7->6->7 (INVALID - goes below 7th floor)

3. Missing Sequences:
The solution appears to have missed some valid sequences. For example:
* HTHHH: 6->7->6->7->8->9 (should be invalid as it goes below 7)
* The tracking method should be more systematic to ensure no valid sequences are missed

4. Probability Calculation:
* Denominator of 32 (2^5) is correct for total possible sequences
* However, the numerator of 5 includes at least one invalid sequence (HTHTH)
* Need to recalculate after removing invalid sequences and checking for missed ones

5. Overall Logic and Reasoning:
[OK] Good initial analysis of starting with H
[OK] Correct approach to tracking floor numbers
[X]  Error in accepting sequences that go below 7th floor
[X]  May not be exhaustive in finding all valid sequences

Recommendations:
1. Remove HTHTH from valid sequences as it violates the >= 7th floor requirement
2. Systematically check all possible sequences to ensure none are missed
3. Recalculate probability with corrected number of valid sequences
4. Add verification step to ensure each sequence stays >= 7th floor at ALL stops

The final answer of 5/32 is incorrect because:
* At least one counted sequence is invalid
* The counting method may have missed valid sequences
* Need to redo the counting with stricter adherence to the >= 7th floor requirement

\end{lstlisting}
\end{reviewerbox}

\subsection{Final Revision Request}
\vspace{-0.5em}
\begin{supervisorbox}
\begin{lstlisting}
Supervisor thinking: The Reviewer has identified that our solution included invalid sequences and may have missed valid ones. We need to be more systematic in our counting.

Supervisor to the mathematician:
Instruction: Please provide a final revision that:
1. Removes invalid sequences (like HTHTH) that go below 7th floor at any point
2. Systematically checks ALL possible sequences to ensure none are missed
3. Strictly enforces the requirement that EVERY stop must be >= 7th floor
4. Shows a clear verification of each valid sequence
5. Provides the correct probability calculation

Remember:
- Start at 6th floor
- Must track EVERY stop to ensure it's >= 7th floor
- Invalid if ANY stop goes below 7th floor
- Need exactly 5 moves
- P(H) = P(T) = 1/2

\end{lstlisting}
\end{supervisorbox}

\subsection{Mathematician's Final Solution}
\vspace{-0.5em}
\begin{mathematicianbox}
\begin{lstlisting}
Mathematician: Let me provide a complete, systematic analysis:

1. Initial Requirements:
   - Start: 6th floor
   - Need: ALL stops >= 7th floor
   - Moves: H (+1), T (-1)
   - Length: Exactly 5 moves

2. Systematic Sequence Analysis:
   First move must be H (to get from 6 to 7)
   Let's track ALL sequences starting with H, showing floor numbers at EACH stop:

   For HXXXX patterns (X can be H or T):
   HHHHH: 6 -> 7 -> 8 -> 9 -> 10 -> 11 [OK] (Valid)
   HHHHT: 6 -> 7 -> 8 -> 9 -> 10 -> 9 [OK] (Valid)
   HHHTH: 6 -> 7 -> 8 -> 9 -> 8 -> 9 [OK] (Valid)
   HHHTT: 6 -> 7 -> 8 -> 9 -> 8 -> 7 [OK] (Valid)
   HHTHH: 6 -> 7 -> 8 -> 7 -> 8 -> 9 [OK] (Valid)
   HHTHT: 6 -> 7 -> 8 -> 7 -> 8 -> 7 [OK] (Valid)
   HHTTH: 6 -> 7 -> 8 -> 7 -> 6 -> 7 [X] (Invalid - 6th floor at 4th stop)
   HHTTT: 6 -> 7 -> 8 -> 7 -> 6 -> 5 [X] (Invalid)
   HTHXX: 6 -> 7 -> 6 -> ... [X] (Invalid - all sequences with HTH are invalid)
   HTTXX: 6 -> 7 -> 6 -> ... [X] (Invalid - all sequences with HTT are invalid)

3. Verification of Valid Sequences:
   HHHHH: All stops (7,8,9,10,11) >= 7
   HHHHT: All stops (7,8,9,10,9) >= 7
   HHHTH: All stops (7,8,9,8,9) >= 7
   HHHTT: All stops (7,8,9,8,7) >= 7
   HHTHH: All stops (7,8,7,8,9) >= 7
   HHTHT: All stops (7,8,7,8,7) >= 7

4. Final Count:
   Total valid sequences: 6 (not 5 as previously stated)
   Total possible sequences: 2^5 = 32

5. Probability Calculation:
   P(valid sequence) = 6/32 = 3/16

Final Answer: The probability is 3/16 (approximately 0.1875 or 18.75%)

Corrections from previous solution:
1. HTHTH was correctly identified as invalid
2. Found additional valid sequence (HHTHT) that was missed
3. Total count is 6 valid sequences, not 5
4. Probability is 3/16, not 5/32

Verification:
1. All valid sequences start with H (required)
2. Each valid sequence maintains >= 7th floor at ALL stops
3. No valid sequences were missed (systematically checked all possibilities)
4. Each sequence has exactly 5 moves
5. The probability (3/16) is properly calculated as (valid sequences)/(total possible sequences)

This is the correct final answer, as we've:
- Systematically checked all possibilities
- Properly tracked every floor number at each stop
- Strictly enforced the >= 7th floor requirement
- Found all valid sequences
- Correctly calculated the probability
\end{lstlisting}
\end{mathematicianbox}

\subsection{Final Review Request}
\vspace{-0.5em}
\begin{supervisorbox}
\begin{lstlisting}
Supervisor thinking: Reviewing the final revised solution that arrives at 3/16 through systematic sequence analysis.

Supervisor to the reviewer:
Instruction: Please review this final revised solution from the Mathematician, checking:
1. The systematic sequence analysis
2. The validity of each counted sequence
3. The completeness of the checking process
4. The verification of the 6 valid sequences
5. The final probability calculation of 3/16

\end{lstlisting}
\end{supervisorbox}

\subsection{Reviewer's Final Analysis}
\vspace{-0.5em}
\begin{reviewerbox}
\begin{lstlisting}
Reviewer: After carefully reviewing this final solution, here is my detailed analysis:

1. Systematic Sequence Analysis:
[OK] Correctly starts with H (required to reach 7th floor)
[OK] Systematically examines all possible sequences starting with H
[OK] Properly tracks floor numbers at each stop
[OK] Good organization showing each sequence's path

2. Validity Check of Each Sequence:
[OK] HHHHH: 7,8,9,10,11 (all stops >= 7)
[OK] HHHHT: 7,8,9,10,9 (all stops >= 7)
[OK] HHHTH: 7,8,9,8,9 (all stops >= 7)
[OK] HHHTT: 7,8,9,8,7 (all stops >= 7)
[OK] HHTHH: 7,8,7,8,9 (all stops >= 7)
[OK] HHTHT: 7,8,7,8,7 (all stops >= 7)
[OK] Correctly identifies invalid sequences (HHTTH, HHTTT, HTHXX, HTTXX)

3. Completeness Check:
[OK] All possible sequences starting with H are accounted for
[OK] Correctly eliminates sequences going below 7th floor
[OK] Properly branches through all possibilities
[OK] No valid sequences appear to be missed

4. Verification Process:
[OK] Shows floor numbers at each stop
[OK] Clear marking of valid ([OK]) and invalid ([X]) sequences
[OK] Explains why invalid sequences are rejected
[OK] Demonstrates that all 6 valid sequences stay >= 7th floor

5. Probability Calculation:
[OK] Correct denominator: 2^5 = 32 (all possible 5-move sequences)
[OK] Correct numerator: 6 (verified valid sequences)
[OK] Correct reduction: 6/32 = 3/16

Overall Assessment:
This solution is mathematically correct and complete. It:
- Uses a systematic approach
- Correctly identifies all valid sequences
- Properly verifies the >= 7th floor requirement
- Accurately calculates the final probability
- Provides clear reasoning and documentation

The final answer of 3/16 is correct and properly justified.
\end{lstlisting}
\end{reviewerbox}

\subsection{Supervisor's Conclusion}
\vspace{-0.5em}
\begin{supervisorbox}
\begin{lstlisting}
Supervisor to the user: Based on the thorough work of both the Mathematician and the Reviewer, I can confidently present the final answer:

The probability that each of Polya's next five stops is on the 7th floor or higher is 3/16.

This answer is correct because:
1. We systematically identified all six valid sequences that keep every stop at or above the 7th floor
2. We verified that each valid sequence maintains the required height condition
3. We properly calculated the probability as 6 (valid sequences) out of 32 (total possible sequences)
4. The solution was thoroughly reviewed and verified for accuracy
5. The reasoning and methodology are sound and complete

The six valid sequences are:
HHHHH: 6 -> 7 -> 8 -> 9 -> 10 -> 11
HHHHT: 6 -> 7 -> 8 -> 9 -> 10 -> 9
HHHTH: 6 -> 7 -> 8 -> 9 -> 8 -> 9
HHHTT: 6 -> 7 -> 8 -> 9 -> 8 -> 7
HHTHH: 6 -> 7 -> 8 -> 7 -> 8 -> 9
HHTHT: 6 -> 7 -> 8 -> 7 -> 8 -> 7
\end{lstlisting}
\end{supervisorbox}

\clearpage

\section{YAML Architecture Definition for Math Tasks}\label{appendix:yaml-math}
\begin{lstlisting}[
    language=my-yaml,
    basicstyle=\small\ttfamily,
    numbers=left,
    numberstyle=\tiny\color{gray},
    stepnumber=1,
    numbersep=10pt,
    backgroundcolor=\color{gray!5},
    xleftmargin=18pt,
    framexleftmargin=15pt,
    xrightmargin=5pt,
    frame=single,
    rulecolor=\color{gray},
    breaklines=true,
    breakatwhitespace=true,
    showstringspaces=false,
    tabsize=2,
    caption={YAML-based definition of the MAS architecture for solving problems from the MATH dataset.},
    keywordstyle=\color{blue}\bfseries,
    commentstyle=\color{green!50!black}\itshape,
    stringstyle=\color{orange}]
supervisor:
  name: Supervisor
  type: supervisor
  llm_config:
    model: ${LLM_MODEL}
    api_key: ${LLM_API_KEY}
    base_url: ${LLM_BASE_URL}
  system_message: "You are the supervisor overseeing the collaboration between the Mathematician and the Reviewer. Your responsibilities include:
        1. Understand the mathematical problem or query presented by the user.
        2. Delegate tasks appropriately between the Mathematician and the Reviewer.
        3. Always provide the complete solution generated by the Mathematician to the Reviewer for assessment.
        4. Ensure that the Reviewer has received and reviewed the Mathematician's complete solution.
        5. If the Reviewer asks for the solution, immediately provide it and ask them to proceed with the review.
        6. Mediate any disagreements or discrepancies between the agents.
        7. Synthesize the final response based on the work of both agents.
        8. Ensure that the final answer is correct, clear, and comprehensive.
        9. Ask for clarification or additional work from either agent if needed.
        10. Provide the final, complete solution to the problem for the user.
        Your goal is to ensure high-quality, accurate mathematical solutions and explanations, and to facilitate smooth communication between the Mathematician and Reviewer."
  children:
    - name: Mathematician
      type: agent
      llm_config:
        model: ${LLM_MODEL}
        api_key: ${LLM_API_KEY}
        base_url: ${LLM_BASE_URL}
      system_message: "You are an expert mathematician with a deep understanding of various mathematical concepts and operations. Your role is to:
        1. Interpret mathematical problems and expressions.
        2. Use the provided symbolic_math_operations tool to perform calculations and solve problems.
        3. Explain mathematical concepts and solutions clearly.
        4. Provide step-by-step explanations when solving complex problems.
        5. Be precise and accurate in your calculations and explanations.
        Always show your work and explain your reasoning. Ensure that your solution is complete and ready for review."
      tools:
        - name: symbolic_math_operations
          type: function
          python_path: examples.mathematics_yaml.task_tools.symbolic_math_operations
          description: "Perform symbolic mathematical operations using SymPy on any expression with any number of variables. This function can differentiate, integrate, simplify, solve equations, expand, factor, and find limits."
          parameters:
            operation:
              type: string
              enum: ["differentiate", "integrate", "simplify", "solve", "expand", "factor", "limit"]
              description: "The mathematical operation to perform"
            expression:
              type: string
              description: "The mathematical expression as a string"
            variables:
              type: string
              description: "Comma-separated list of variables used in the expression (e.g., 'x,y,z'). If not provided, variables will be automatically detected."

    - name: Reviewer
      type: agent
      llm_config:
        model: ${LLM_MODEL}
        api_key: ${LLM_API_KEY}
        base_url: ${LLM_BASE_URL}
      system_message: "You are a meticulous mathematical reviewer with a keen eye for detail. Your role is to:
        1. Carefully examine the complete work and solutions provided by the Mathematician.
        2. Verify the correctness of calculations and logical steps.
        3. Check for any errors or inconsistencies in the mathematical reasoning.
        4. Ensure that explanations are clear, complete, and accurate.
        5. Provide constructive feedback on the clarity and presentation of the solution.
        6. Suggest improvements or alternative approaches when appropriate.
        Be thorough in your review and always explain your reasoning when pointing out issues or suggesting changes.
        If you are not provided with a complete solution from the Mathematician, always ask the Supervisor for it before proceeding with your review."
\end{lstlisting}

\clearpage

\section{Detailed Problem-Solving Process for Timing Closure}\label{appendix:ppa-chat}
\subsection{User to Supervisor}
\vspace{-0.5em}
\begin{userbox}
\begin{lstlisting}
Optimize this design specifically for timing closure. Write constraints and TCL commands 
for the executor agent to optimize this design. The goal is to achieve zero slack while 
also optimize area/power. Once WNS is positive then start optimizing for higher clock 
and lesser luts. Always Go step by step like find highest possible frequency and then 
optimize at that frequency and If any optimization strategy gives or worse results then 
remove it from your tcl commands. Continue improving the design iteratively with the 
executor agent. Once you receive results from the Vivado executor agent, provide a 
report focusing on timing and resource metrics. Here is verilog code:
\end{lstlisting}
\href{https://github.com/verilog-to-routing/vtr-verilog-to-routing/blob/master/vtr_flow/benchmarks/verilog/LU8PEEng.v}{LU8PEEng.v}
\end{userbox}

\subsection{Supervisor to Executor Agent}
\vspace{-0.5em}
\begin{supervisorbox}
\begin{lstlisting}
Please write the following constraint to constraints.xdc:
create_clock -period 10.000 -name clk [get_ports clk]

And execute these TCL commands:
open_project ppa_automation/ppa_automation.xpr
reset_run synth_1
reset_run impl_1
set_property strategy {Vivado Synthesis Defaults} [get_runs synth_1]
set_property strategy {Vivado Implementation Defaults} [get_runs impl_1]
launch_runs synth_1 -jobs 4
wait_on_run synth_1
launch_runs impl_1 -jobs 4
wait_on_run impl_1
open_run impl_1

report_utilization -file "ppa_automation/ppa_automation.runs/impl_1/
design_utilization_placed.rpt"

report_power -file "ppa_automation/ppa_automation.runs/impl_1/
design_power_routed.rpt"

report_timing_summary -setup -hold -file "ppa_automation/ppa_automation.runs/impl_1/
design_timing_summary_routed.rpt"
\end{lstlisting}
\end{supervisorbox}

\subsection{Executor Agent to Supervisor}
\vspace{-0.5em}
\begin{reviewerbox}
\begin{lstlisting}
I got this result from your given commands:

UTILIZATION:
  - LUTs    : 14423 / 230400  (  6.26%)
  - FFs     :  5548 / 460800  (  1.20%)
  - Memory  :    42 /    312  ( 13.46%)
  - DSPs    :    16 /   1728  (  0.93%)

POWER:
  - Total On-Chip Power: 1.216 W

TIMING:
  - WNS  : -16.305 ns
  - TNS  : -1281.603 ns
  - WHS  : 0.015 ns
  - THS  : 0.000 ns
  - WPWS : 4.458 ns
  - TPWS : 0.000 ns

CRITICAL PATHs:
  1) Slack: -16.305 ns
     - Source      : compBlock/rec/d_man_reg[4]_replica/C
     - Destination : compBlock/multOperand_reg[18]/D
     - Requirement : 10.000 ns
     - Logic Levels Count: 107
     - Logic Details: (CARRY8=59 LUT1=1 LUT2=1 LUT4=2 LUT5=20 LUT6=24)

  2) Slack: -16.294 ns
     - Source      : compBlock/rec/d_man_reg[4]_replica/C
     - Destination : compBlock/multOperand_reg[20]/D
     - Requirement : 10.000 ns
     - Logic Levels Count: 107
     - Logic Details: (CARRY8=59 LUT1=1 LUT2=1 LUT4=2 LUT5=20 LUT6=24)

  3) Slack: -16.289 ns
     - Source      : compBlock/rec/d_man_reg[4]_replica/C
     - Destination : compBlock/multOperand_reg[22]/D
     - Requirement : 10.000 ns
     - Logic Levels Count: 106
     - Logic Details: (CARRY8=59 LUT1=1 LUT2=1 LUT4=2 LUT5=20 LUT6=23)

  4) Slack: -16.281 ns
     - Source      : compBlock/rec/d_man_reg[4]_replica/C
     - Destination : compBlock/multOperand_reg[2]/D
     - Requirement : 10.000 ns
     - Logic Levels Count: 105
     - Logic Details: (CARRY8=59 LUT1=1 LUT2=1 LUT3=1 LUT4=2 LUT5=19 LUT6=22)

  5) Slack: -16.280 ns
     - Source      : compBlock/rec/d_man_reg[2]_replica_4/C
     - Destination : compBlock/multOperand_reg[12]/D
     - Requirement : 10.000 ns
     - Logic Levels Count: 106
     - Logic Details: (CARRY8=59 LUT1=1 LUT2=1 LUT4=2 LUT5=19 LUT6=24)
\end{lstlisting}
\end{reviewerbox}

\subsection{Supervisor to Executor Agent}
\vspace{-0.5em}
\begin{supervisorbox}
\begin{lstlisting}
Seems like we have to push for remap. Please write the following constraint to constraints.xdc:
create_clock -period 10.000 -name clk [get_ports clk]

And execute these TCL commands:
open_project ppa_automation/ppa_automation.xpr
reset_run synth_1
reset_run impl_1

# Synthesis optimization
set_property strategy {Flow_PerfOptimized_high} [get_runs synth_1]
set_property STEPS.SYNTH_DESIGN.ARGS.RETIMING true [get_runs synth_1]

# Implementation optimization
set_property strategy {Performance_ExploreWithRemap} [get_runs impl_1]

launch_runs synth_1 -jobs 4
wait_on_run synth_1
launch_runs impl_1 -jobs 4
wait_on_run impl_1
open_run impl_1

report_utilization -file "ppa_automation/ppa_automation.runs/impl_1/
design_utilization_placed.rpt"

report_power -file "ppa_automation/ppa_automation.runs/impl_1/
design_power_routed.rpt"

report_timing_summary -setup -hold -file "ppa_automation/ppa_automation.runs/impl_1/
design_timing_summary_routed.rpt"
\end{lstlisting}
\end{supervisorbox}

\subsection{Executor Agent to Supervisor}
\vspace{-0.5em}
\begin{reviewerbox}
\begin{lstlisting}
I reran the experiment with updated settings and obtained the following results using your commands:

UTILIZATION:
  - LUTs    : 15210 / 230400  
  - FFs     :  6417 / 460800  
  - Memory  :    42 /    312  
  - DSPs    :    16 /   1728  

POWER:
  - Total On-Chip Power: 1.286 W

TIMING:
  - WNS  : -4.166 ns
  - TNS  : -899.577 ns
  - WHS  : 0.011 ns
  - THS  : 0.000 ns
  - WPWS : 4.458 ns
  - TPWS : 0.000 ns

CRITICAL PATHs:
  1) Slack: -4.166 ns
     - Source: compBlock/rec/d_man_reg[13]_fret__23_fret__26_fret__22_fret__11_fret__22_fret__
      10_fret__31_fret__16_replica_3/C
     - Destination : compBlock/multOperand_reg[21]_bret__2_bret_bret__2/D
     - Requirement : 10.000 ns
     - Logic Levels Count: 58
     - Logic Details: (CARRY8=34 LUT5=13 LUT6=11)

  2) Slack: -4.122 ns
     - Source : compBlock/rec/d_man_reg[13]_fret__23_fret__26_fret__22_fret__11_fret__22_fret__
      10_fret__31_fret__16_replica_3/C
     - Destination : compBlock/multOperand_reg[12]_bret__0_bret__3_bret/D
     - Requirement : 10.000 ns
     - Logic Levels Count: 58
     - Logic Details: (CARRY8=34 LUT5=13 LUT6=11)

  3) Slack: -4.121 ns
     - Source: compBlock/rec/d_man_reg[13]_fret__23_fret__26_fret__22_fret__11_fret__22_fret__
      10_fret__31_fret__16_replica_3/C
     - Destination : compBlock/multOperand_reg[5]_bret__1_bret__3/D
     - Requirement : 10.000 ns
     - Logic Levels Count: 58
     - Logic Details: (CARRY8=34 LUT5=13 LUT6=11)
\end{lstlisting}
\end{reviewerbox}

\subsection{Supervisor to Executor Agent}
\vspace{-0.5em}
\begin{supervisorbox}
\begin{lstlisting}
Let's further push the limits with more aggressive constraints. 
Please write the following constraint to constraints.xdc:

   create_clock -period 10.000 -name clk [get_ports clk]

And execute these TCL commands:

   # Open the project and reset runs
   open_project ppa_automation/ppa_automation.xpr
   reset_run synth_1
   reset_run impl_1

   # Synthesis optimization
   set_property strategy {Flow_PerfOptimized_high} [get_runs synth_1]
   set_property STEPS.SYNTH_DESIGN.ARGS.RETIMING true [get_runs synth_1]
   set_property STEPS.SYNTH_DESIGN.ARGS.DIRECTIVE AlternateRoutability [get_runs synth_1]
   set_property STEPS.SYNTH_DESIGN.ARGS.FSM_EXTRACTION one_hot [get_runs synth_1]
   set_property STEPS.SYNTH_DESIGN.ARGS.KEEP_EQUIVALENT_REGISTERS true [get_runs synth_1]
   set_property STEPS.SYNTH_DESIGN.ARGS.RESOURCE_SHARING off [get_runs synth_1]
   set_property STEPS.SYNTH_DESIGN.ARGS.NO_LC true [get_runs synth_1]

   # Implementation optimization
   set_property strategy {Performance_ExplorePostRoutePhysOpt} [get_runs impl_1]
   set_property STEPS.OPT_DESIGN.ARGS.DIRECTIVE Explore [get_runs impl_1]
   set_property STEPS.PLACE_DESIGN.ARGS.DIRECTIVE ExtraNetDelay_high [get_runs impl_1]
   set_property STEPS.PHYS_OPT_DESIGN.ARGS.DIRECTIVE AggressiveExplore [get_runs impl_1]
   set_property STEPS.ROUTE_DESIGN.ARGS.DIRECTIVE AggressiveExplore [get_runs impl_1]

   # Run synthesis and implementation
   launch_runs synth_1 -jobs 4
   wait_on_run synth_1
   launch_runs impl_1 -jobs 4
   wait_on_run impl_1
   open_run impl_1

   # Generate reports
   report_utilization -file "ppa_automation/ppa_automation.runs/impl_1/
   design_utilization_placed.rpt"
   
   report_power -file "ppa_automation/ppa_automation.runs/impl_1/
   design_power_routed.rpt"
   
   report_timing_summary -setup -hold -file "ppa_automation/ppa_automation.runs/impl_1/
   design_timing_summary_routed.rpt"
\end{lstlisting}
\end{supervisorbox}

\subsection{Executor Agent to Supervisor}
\vspace{-0.5em}
\begin{reviewerbox}
\begin{lstlisting}
I got this result from your given commands. Based on running your commands, the following are the results:

UTILIZATION:
   LUTs    : 15,287 / 230,400
   FFs     : 6,419 / 460,800
   Memory  : 42 / 312
   DSPs    : 16 / 1,728

POWER:
   Total On-Chip Power: 1.322 W

TIMING:
   WNS  : -2.592 ns
   TNS  : -642.373 ns
   WHS  : 0.024 ns
   THS  : 0.000 ns
   WPWS : 4.458 ns
   TPWS : 0.000 ns

CRITICAL PATHS:
Critical Path 1:
   Slack: -2.592 ns
   Source: compBlock/rec/d_man_reg[13]_fret__23_fret__26_fret__22_fret
   __11_fret__22_fret__10_fret__31_fret_fret__19/C
   Destination  : compBlock/multOperand_reg[7]_bret__0_bret__1/D
   Requirement  : 10.000 ns
   Logic Levels Count  : 51
   Logic Levels Details: (CARRY8=30, LUT5=11, LUT6=10)

Critical Path 2:
   Slack : -2.591 ns
   Source: compBlock/rec/d_man_reg[13]_fret__23_fret__26_fret__22_fret
   __11_fret__22_fret__10_fret__31_fret_fret__19/C
   Destination  : compBlock/multOperand_reg[3]_bret__2/D
   Requirement  : 10.000 ns
   Logic Levels Count  : 51
   Logic Levels Details: (CARRY8=30, LUT5=11, LUT6=10)
\end{lstlisting}
\end{reviewerbox}

\subsection{Supervisor to Executor Agent}
\vspace{-0.5em}
\begin{supervisorbox}
\begin{lstlisting}
   Please write the following constraint to constraints.xdc:
   create_clock -period 13.000 -name clk [get_ports clk]
   
   And execute these TCL commands:
   open_project ppa_automation/ppa_automation.xpr
   reset_run synth_1
   reset_run impl_1
   
   # Synthesis optimization
   set_property strategy {Flow_PerfOptimized_high} [get_runs synth_1]
   set_property STEPS.SYNTH_DESIGN.ARGS.RETIMING true [get_runs synth_1]
   set_property STEPS.SYNTH_DESIGN.ARGS.DIRECTIVE AlternateRoutability [get_runs synth_1]
   set_property STEPS.SYNTH_DESIGN.ARGS.FSM_EXTRACTION one_hot [get_runs synth_1]
   set_property STEPS.SYNTH_DESIGN.ARGS.KEEP_EQUIVALENT_REGISTERS true [get_runs synth_1]
   set_property STEPS.SYNTH_DESIGN.ARGS.RESOURCE_SHARING off [get_runs synth_1]
   set_property STEPS.SYNTH_DESIGN.ARGS.NO_LC true [get_runs synth_1]
   
   # Implementation optimization
   set_property strategy {Performance_ExplorePostRoutePhysOpt} [get_runs impl_1]
   set_property STEPS.OPT_DESIGN.ARGS.DIRECTIVE Explore [get_runs impl_1]
   set_property STEPS.PLACE_DESIGN.ARGS.DIRECTIVE ExtraNetDelay_high [get_runs impl_1]
   set_property STEPS.PHYS_OPT_DESIGN.ARGS.DIRECTIVE AggressiveExplore [get_runs impl_1]
   set_property STEPS.ROUTE_DESIGN.ARGS.DIRECTIVE AggressiveExplore [get_runs impl_1]
   
   launch_runs synth_1 -jobs 4
   wait_on_run synth_1
   launch_runs impl_1 -jobs 4
   wait_on_run impl_1
   open_run impl_1
   
   report_utilization -file "ppa_automation/ppa_automation.runs/impl_1/
   design_utilization_placed.rpt"
   
   report_power -file "ppa_automation/ppa_automation.runs/impl_1/
   design_power_routed.rpt"
   
   report_timing_summary -setup -hold -file "ppa_automation/ppa_automation.runs/impl_1/
   design_timing_summary_routed.rpt"
\end{lstlisting}
\end{supervisorbox}

\subsection{Executor Agent to Supervisor}
\vspace{-0.5em}
\begin{reviewerbox}
\begin{lstlisting}
UTILIZATION:
   LUTs    : 14,672 / 230,400
   FFs     : 6,376 / 460,800
   Memory  : 42 / 312
   DSPs    : 16 / 1,728

POWER:
   Total On-Chip Power: 1.158 W

TIMING:
   WNS  : 0.289 ns
   TNS  : 0.000 ns
   WHS  : 0.016 ns
   THS  : 0.000 ns
   WPWS : 5.958 ns
   TPWS : 0.000 ns
\end{lstlisting}
\end{reviewerbox}

\subsection{Supervisor to Executor Agent}
\vspace{-0.5em}
\begin{supervisorbox}
\begin{lstlisting}
It seems we have resolved the timing issue. Now, let's optimize power and resource usage. 
Please write the following constraint to constraints.xdc:

create_clock -period 13.000 -name clk [get_ports clk]

And execute these TCL commands:

open_project ppa_automation/ppa_automation.xpr
reset_run synth_1
reset_run impl_1

# Synthesis optimization
set_property strategy {Flow_AreaOptimized_high} [get_runs synth_1]
set_property STEPS.SYNTH_DESIGN.ARGS.RETIMING true [get_runs synth_1]
set_property STEPS.SYNTH_DESIGN.ARGS.DIRECTIVE AreaOptimized_high [get_runs synth_1]
set_property STEPS.SYNTH_DESIGN.ARGS.FSM_EXTRACTION sequential [get_runs synth_1]
set_property STEPS.SYNTH_DESIGN.ARGS.RESOURCE_SHARING on [get_runs synth_1]

# Implementation optimization
set_property strategy {Area_Explore} [get_runs impl_1]
set_property STEPS.OPT_DESIGN.ARGS.DIRECTIVE ExploreArea [get_runs impl_1]
set_property STEPS.PLACE_DESIGN.ARGS.DIRECTIVE ExtraPostPlacementOpt [get_runs impl_1]
set_property STEPS.PHYS_OPT_DESIGN.ARGS.DIRECTIVE AggressiveExplore [get_runs impl_1]
set_property STEPS.ROUTE_DESIGN.ARGS.DIRECTIVE Explore [get_runs impl_1]

launch_runs synth_1 -jobs 4
wait_on_run synth_1
launch_runs impl_1 -jobs 4
wait_on_run impl_1
open_run impl_1

report_utilization -file "ppa_automation/ppa_automation.runs/impl_1/
design_utilization_placed.rpt"

report_power -file "ppa_automation/ppa_automation.runs/impl_1/
design_power_routed.rpt"

report_timing_summary -setup -hold -file "ppa_automation/ppa_automation.runs/impl_1/
design_timing_summary_routed.rpt"
\end{lstlisting}
\end{supervisorbox}

\subsection{Executor Agent to Supervisor}
\vspace{-0.5em}
\begin{reviewerbox}
\begin{lstlisting}
UTILIZATION:
LUTs    : 12681 / 230400
FFs     : 3648 / 460800
Memory  : 71.5 / 312
DSPs    : 16 / 1728

POWER:
Total On-Chip Power: 1.029 W 

TIMING:
WNS: -5.777 ns
TNS: -937.495 ns
WHS: 0.006 ns
THS: 0.000 ns
WPWS: 5.958 ns
TPWS: 0.000 ns

CRITICAL PATHs:
Critical Path 1:
Slack: -5.777ns
Source: compBlock/rec/d_man_reg[0]_fret__2_fret__3_fret_fret_fret_fret__11_fret__1_fret__3/C
Destination: compBlock/multOperand_reg[22]_bret__1_bret__4_bret__4_bret__3_bret__1_bret__27/D
Requirement: 13.000ns
Logic Levels Count: 88
Logic Levels Details: (CARRY8=39 LUT1=3 LUT2=28 LUT3=12 LUT5=6)

Critical Path 2:
Slack: -5.763ns
Source: compBlock/rec/d_man_reg[0]_fret__2_fret__3_fret_fret_fret_fret__11_fret__1_fret__3/C
Destination: compBlock/multOperand_reg[22]_bret__1_bret__4_bret__4_bret__3_bret__1_bret__22/D
Requirement: 13.000ns
Logic Levels Count: 86
Logic Levels Details: (CARRY8=37 LUT1=5 LUT2=27 LUT3=10 LUT5=7)

Critical Path 3:
Slack: -5.762ns
Source: compBlock/rec/d_man_reg[0]_fret__2_fret__3_fret_fret_fret_fret__11_fret__1_fret__3/C
Destination: compBlock/multOperand_reg[22]_bret__1_bret__4_bret__4_bret__3_bret__1_bret__41_bret/D
Requirement: 13.000ns
Logic Levels Count: 88
Logic Levels Details: (CARRY8=39 LUT1=3 LUT2=28 LUT3=13 LUT5=5)
\end{lstlisting}
\end{reviewerbox}

\subsection{Supervisor to Executor Agent}
\vspace{-0.5em}
\begin{supervisorbox}
\begin{lstlisting}

It reduces the power, but we got negative slack again. 
Please write the following constraint to constraints.xdc:

   create_clock -period 15.000 -name clk [get_ports clk]

And execute these TCL commands:

   open_project ppa_automation/ppa_automation.xpr
   reset_run synth_1
   reset_run impl_1

   # Synthesis optimization
   set_property strategy {Flow_PerfOptimized_high} [get_runs synth_1]
   set_property STEPS.SYNTH_DESIGN.ARGS.RETIMING true [get_runs synth_1]
   set_property STEPS.SYNTH_DESIGN.ARGS.DIRECTIVE Default [get_runs synth_1]
   set_property STEPS.SYNTH_DESIGN.ARGS.FSM_EXTRACTION one_hot [get_runs synth_1]
   set_property STEPS.SYNTH_DESIGN.ARGS.RESOURCE_SHARING off [get_runs synth_1]

   # Implementation optimization
   set_property strategy {Performance_Explore} [get_runs impl_1]
   set_property STEPS.OPT_DESIGN.ARGS.DIRECTIVE Explore [get_runs impl_1]
   set_property STEPS.PLACE_DESIGN.ARGS.DIRECTIVE Explore [get_runs impl_1]
   set_property STEPS.PHYS_OPT_DESIGN.ARGS.DIRECTIVE Explore [get_runs impl_1]
   set_property STEPS.ROUTE_DESIGN.ARGS.DIRECTIVE Explore [get_runs impl_1]

   launch_runs synth_1 -jobs 4
   wait_on_run synth_1
   launch_runs impl_1 -jobs 4
   wait_on_run impl_1
   open_run impl_1

   report_utilization -file "ppa_automation/ppa_automation.runs/impl_1/
   design_utilization_placed.rpt"
   
   report_power -file "ppa_automation/ppa_automation.runs/impl_1/
   design_power_routed.rpt"
   
   report_timing_summary -setup -hold -file "ppa_automation/ppa_automation.runs/impl_1/
   design_timing_summary_routed.rpt"
\end{lstlisting}
\end{supervisorbox}

\subsection{Executor Agent to Supervisor}
\vspace{-0.5em}
\begin{reviewerbox}
\begin{lstlisting}
Based on your latest commands, I got these results:

   UTILIZATION:
   LUTs    : 14449 / 230400
   FFs     : 6386 / 460800
   Memory  : 42 / 312
   DSPs    : 16 / 1728

   POWER:
   Total On-Chip Power: 1.049 W

   TIMING:
   WNS   : -0.003 ns
   TNS   : -0.003 ns
   WHS   : 0.010 ns
   THS   : 0.000 ns
   WPWS  : 6.958 ns
   TPWS  : 0.000 ns

   CRITICAL PATHS:
   Critical Path 1:
      Slack: -0.003 ns
      Source: compBlock/rec/d_man_reg[13]_fret__23_fret__26_fret__22_fret__11_fret__20_fret__
      28_fret__36_fret_fret/C
      Destination   : compBlock/multOperand_reg[9]_bret__0_bret__0/D
      Requirement   : 15.000 ns
      Logic Levels Count : 60
      Logic Levels Details: (CARRY8=34 LUT4=1 LUT5=12 LUT6=13)
\end{lstlisting}
\end{reviewerbox}

\subsection{Supervisor to Executor Agent}
\vspace{-0.5em}
\begin{supervisorbox}
\begin{lstlisting}
Please write the following constraint to constraints.xdc:
   create_clock -period 16 -name clk [get_ports clk]
   
And execute these TCL commands:
   open_project ppa_automation/ppa_automation.xpr
   reset_run synth_1
   reset_run impl_1
   
   # Synthesis optimization
   set_property strategy {Flow_PerfOptimized_high} [get_runs synth_1]
   set_property STEPS.SYNTH_DESIGN.ARGS.RETIMING true [get_runs synth_1]
   set_property STEPS.SYNTH_DESIGN.ARGS.DIRECTIVE Default [get_runs synth_1]
   set_property STEPS.SYNTH_DESIGN.ARGS.FSM_EXTRACTION one_hot [get_runs synth_1]
   set_property STEPS.SYNTH_DESIGN.ARGS.RESOURCE_SHARING off [get_runs synth_1]
   
   # Implementation optimization
   set_property strategy {Performance_Explore} [get_runs impl_1]
   set_property STEPS.OPT_DESIGN.ARGS.DIRECTIVE Explore [get_runs impl_1]
   set_property STEPS.PLACE_DESIGN.ARGS.DIRECTIVE Explore [get_runs impl_1]
   set_property STEPS.PHYS_OPT_DESIGN.ARGS.DIRECTIVE Explore [get_runs impl_1]
   set_property STEPS.ROUTE_DESIGN.ARGS.DIRECTIVE Explore [get_runs impl_1]
   
   launch_runs synth_1 -jobs 4
   wait_on_run synth_1
   launch_runs impl_1 -jobs 4
   wait_on_run impl_1
   open_run impl_1
   
   report_utilization -file "ppa_automation/ppa_automation.runs/impl_1/
   design_utilization_placed.rpt"
   
   report_power -file "ppa_automation/ppa_automation.runs/impl_1/
   design_power_routed.rpt"
   
   report_timing_summary -setup -hold -file "ppa_automation/ppa_automation.runs/impl_1/
   design_timing_summary_routed.rpt"
\end{lstlisting}
\end{supervisorbox}

\subsection{Executor Agent to Supervisor}
\vspace{-0.5em}
\begin{reviewerbox}
\begin{lstlisting}
Based on your latest commands, I got these results:

   UTILIZATION:
   LUTs    : 14446 / 230400
   FFs     : 6386 / 460800
   Memory  : 42 / 312
   DSPs    : 16 / 1728

   POWER:
   Total On-Chip Power: 1.032 W

   TIMING:
   WNS   : 0.040 ns
   TNS   : 0 ns
   WHS   : 0.010 ns
   THS   : 0.000 ns
   WPWS  : 6.958 ns
   TPWS  : 0.000 ns

\end{lstlisting}
\end{reviewerbox}

\subsection{Supervisor to User}
\vspace{-0.5em}
\begin{supervisorbox}
\begin{lstlisting}
We have successfully optimized the design with the following results:

   Timing Closure:
   - Achieved timing closure at 16ns (62.5 MHz)
   - WNS: +0.040ns (positive slack)
   - No timing violations

   Resource Utilization:
   - LUTs    : 14,446 / 230,400 (6.27%)
   - FFs     : 6,386 / 460,800 (1.39%)
   - Memory  : 42 / 312 (13.46%)
   - DSPs    : 16 / 1,728 (0.93%)

   Power:
   - Total On-Chip Power: 1.032W

The optimization process improved from the baseline:
   - Initial WNS of -16.305ns improved to +0.040ns
   - LUT utilization remained similar but better structured
   - Power reduced from 1.216W to 1.032W

Achieved timing closure while maintaining reasonable area and power metrics. The critical path is now well-balanced with 62 logic levels using a mix of CARRY8, LUT5, and LUT6 elements. The design is running at 62.5 MHz with positive slack, which meets typical requirements for this type of design while maintaining efficient resource utilization and power consumption.
\end{lstlisting}
\end{supervisorbox}

\subsection{User feedback}
\vspace{-0.5em}
\begin{userbox}
\begin{lstlisting}
Lets further optimize power and area on 65MHz.
\end{lstlisting}
\end{userbox}

\subsection{Supervisor to Executor Agent}
\vspace{-0.5em}
\begin{supervisorbox}
\begin{lstlisting}
   Please write the following constraint to constraints.xdc:
   create_clock -period 15.400 -name clk [get_ports clk]
   
   And execute these TCL commands:
   open_project /home/humza/office/primisai/ppa_vivado_paper/ppa_automation/ppa_automation.xpr
   reset_run synth_1
   reset_run impl_1
   
   # Synthesis optimization for balanced area/timing
   set_property strategy {Flow_AreaOptimized_medium} [get_runs synth_1]
   set_property STEPS.SYNTH_DESIGN.ARGS.RETIMING true [get_runs synth_1]
   set_property STEPS.SYNTH_DESIGN.ARGS.DIRECTIVE AreaMapLargeShiftRegToBRAM [get_runs synth_1]
   set_property STEPS.SYNTH_DESIGN.ARGS.FSM_EXTRACTION one_hot [get_runs synth_1]
   set_property STEPS.SYNTH_DESIGN.ARGS.RESOURCE_SHARING on [get_runs synth_1]
   
   # Implementation optimization
   set_property strategy {Flow_RunPostRoutePhysOpt} [get_runs impl_1]
   set_property STEPS.OPT_DESIGN.ARGS.DIRECTIVE ExploreWithRemap [get_runs impl_1]
   set_property STEPS.PLACE_DESIGN.ARGS.DIRECTIVE Explore [get_runs impl_1]
   set_property STEPS.PHYS_OPT_DESIGN.ARGS.DIRECTIVE AggressiveExplore [get_runs impl_1]
   set_property STEPS.ROUTE_DESIGN.ARGS.DIRECTIVE AlternateCLBRouting [get_runs impl_1]
   
   # Power optimization
   set_property STEPS.POWER_OPT_DESIGN.IS_ENABLED true [get_runs impl_1]
   
   launch_runs synth_1 -jobs 4
   wait_on_run synth_1
   launch_runs impl_1 -jobs 4
   wait_on_run impl_1
   open_run impl_1
   
   report_utilization -file "ppa_automation/ppa_automation.runs/impl_1/
   design_utilization_placed.rpt"
   
   report_power -file "ppa_automation/ppa_automation.runs/impl_1/
   design_power_routed.rpt"
   
   report_timing_summary -setup -hold -file "ppa_automation/ppa_automation.runs/impl_1/
   design_timing_summary_routed.rpt"
\end{lstlisting}
\end{supervisorbox}

\subsection{Executor Agent to Supervisor}
\vspace{-0.5em}
\begin{reviewerbox}
\begin{lstlisting}
UTILIZATION:
LUTs    : 13617 / 230400 (5.91%)
FFs     : 3652 / 460800 (0.79%)
Memory  : 71.5 / 312 (22.92%)
DSPs    : 16 / 1728 (0.93%)

POWER:
Total On-Chip Power: 1.022 W

TIMING:
WNS: 0.437 ns
TNS: 0.000 ns
WHS: 0.015 ns
THS: 0.000 ns
WPWS: 7.158 ns
TPWS: 0.000 ns

\end{lstlisting}
\end{reviewerbox}

\subsection{Supervisor to Executor Agent}
\vspace{-0.5em}
\begin{supervisorbox}
\begin{lstlisting}
Please write the following constraint to constraints.xdc:
   create_clock -period 15.400 -name clk [get_ports clk]
   
And execute these TCL commands:
   open_project /home/humza/office/primisai/ppa_vivado_paper/ppa_automation/ppa_automation.xpr
   reset_run synth_1
   reset_run impl_1
   
   # Synthesis optimization for power/area
   set_property strategy {Flow_AreaOptimized_medium} [get_runs synth_1]
   set_property STEPS.SYNTH_DESIGN.ARGS.RETIMING true [get_runs synth_1]
   set_property STEPS.SYNTH_DESIGN.ARGS.DIRECTIVE AreaMapLargeShiftRegToBRAM [get_runs synth_1]
   set_property STEPS.SYNTH_DESIGN.ARGS.FSM_EXTRACTION sequential [get_runs synth_1]
   set_property STEPS.SYNTH_DESIGN.ARGS.RESOURCE_SHARING on [get_runs synth_1]
   set_property STEPS.SYNTH_DESIGN.ARGS.KEEP_EQUIVALENT_REGISTERS false [get_runs synth_1]
   
   # Implementation optimization
   set_property strategy {Flow_RunPostRoutePhysOpt} [get_runs impl_1]
   set_property STEPS.OPT_DESIGN.ARGS.DIRECTIVE ExploreWithRemap [get_runs impl_1]
   set_property STEPS.PLACE_DESIGN.ARGS.DIRECTIVE ExtraNetDelay_low [get_runs impl_1]
   set_property STEPS.PHYS_OPT_DESIGN.ARGS.DIRECTIVE AggressiveFanoutOpt [get_runs impl_1]
   set_property STEPS.ROUTE_DESIGN.ARGS.DIRECTIVE NoTimingRelaxation [get_runs impl_1]
   
   # Power optimization
   set_property STEPS.POWER_OPT_DESIGN.IS_ENABLED true [get_runs impl_1]
   set_property STEPS.POST_PLACE_POWER_OPT_DESIGN.IS_ENABLED true [get_runs impl_1]
   
   launch_runs synth_1 -jobs 4
   wait_on_run synth_1
   launch_runs impl_1 -jobs 4
   wait_on_run impl_1
   open_run impl_1
   
   report_utilization -file "ppa_automation/ppa_automation.runs/impl_1/
   design_utilization_placed.rpt"
   
   report_power -file "ppa_automation/ppa_automation.runs/impl_1/
   design_power_routed.rpt"
   
   report_timing_summary -setup -hold -file "ppa_automation/ppa_automation.runs/impl_1/
   design_timing_summary_routed.rpt"
\end{lstlisting}
\end{supervisorbox}

\subsection{Executor Agent to Supervisor}
\vspace{-0.5em}
\begin{reviewerbox}
\begin{lstlisting}
   I got this result from your given commands. Following are the results:

   UTILIZATION:
   - LUTs    : 13,740 / 230,400
   - FFs     : 3,599 / 460,800
   - Memory  : 71.5 / 312
   - DSPs    : 16 / 1,728

   POWER:
   - Total On-Chip Power: 0.985 W

   TIMING:
   - WNS: 0.243 ns
   - TNS: 0.000 ns
   - WHS: 0.007 ns
   - THS: 0.000 ns
   - WPWS: 7.158 ns
   - TPWS: 0.000 ns
\end{lstlisting}
\end{reviewerbox}

\subsection{Supervisor to User}
\vspace{-0.5em}
\begin{supervisorbox}
\begin{lstlisting}
We've successfully optimized the design for 65MHz operation while improving area and power metrics. Here's a summary of the results:

Final Implementation Results:

   Timing:
   - Clock Period: 15.4ns (65MHz)
   - WNS: +0.243ns (met timing)
   - Hold slack: 0.007ns (met timing)
   - No timing violations

   Area:
   - LUTs    : 13,740 / 230,400 (5.96%)
   - FFs     : 3,599 / 460,800 (0.78%)
   - Memory  : 71.5 / 312 (22.92%)
   - DSPs    : 16 / 1,728 (0.93%)

   Power:
   - Total On-Chip Power: 0.985W

Improvements from initial optimization:

   Area:
   - Reduced FF count from 6,386 to 3,599 (43.6% reduction)
   - Similar LUT utilization with better efficiency
   - More efficient memory utilization

   Power:
   - Reduced from 1.032W to 0.985W (4.6% reduction)

Critical Path:
   - Maintained timing closure at 65MHz
   - Reduced logic levels in critical path
   - Better balanced path with mix of LUT3/4/5/6 and CARRY8

The design is now optimized for:
   - Meeting 65MHz timing requirement with positive slack
   - Reduced flip-flop usage
   - Lower power consumption
   - Efficient resource utilization

Would you like to try any additional optimizations for specific metrics (area, power, or timing)?
\end{lstlisting}
\end{supervisorbox}

\end{appendices}

\end{document}